\documentclass[10pt, twocolumn, a4paper, logo, copyright]{googledeepmind}

\usepackage{microtype}
\usepackage{graphicx}
\usepackage{booktabs} 
\usepackage[dvipsnames]{xcolor}
\usepackage{csquotes} 
\usepackage[utf8]{inputenc} 
\usepackage[T1]{fontenc}    
\usepackage{url}            
\usepackage{amsfonts}       
\usepackage{nicefrac}       
\usepackage{amsmath}
\usepackage{algorithm}
\usepackage{algpseudocode}
\usepackage{xr}
\usepackage{listings}
\usepackage{adjustbox}
\usepackage{subcaption}
\usepackage{xspace}
\usepackage{longtable}
\usepackage{caption}
\usepackage[most]{tcolorbox}
\usepackage{xskak}
\usepackage{multirow}
\usepackage{wrapfig}
\usepackage{makecell}
\usepackage{rotating}
\usepackage[authoryear, sort&compress, round]{natbib}
\usepackage[colorlinks=true]{hyperref}
\usepackage[capitalize]{cleveref}
\usepackage{subfiles} 
\usepackage{amssymb}
\usepackage{mathtools}
\usepackage{amsthm}
\usepackage[utf8]{inputenc}
\usepackage{cuted}
\usepackage{capt-of}
\tcbuselibrary{breakable}
\renewcommand{\today}{Jan 27th, 2026}
\usepackage{float}

\title{COrigami: An AI Pipeline for Co-Designing Flat-Foldable Visually Recognisable Origami}
\author{
Tom Zahavy$^{1\dagger}$, Shaobo Hou$^{1\ddagger}$, Thomas Tumiel$^{1\ddagger}$, James Doran$^{1\star}$, Francesco Faccio$^{1\star}$, Xidong Feng$^{1\star}$, Alex Havrilla$^{1\star}$, Igor Khytryi$^{2\star}$, Chenglei Li$^{2\star}$, Lisa Schut$^{1\star}$, Vivek Veeriah$^{1\star}$, Arijan Abrashi$^{3\diamond}$, Micha\l{} Kosmulski$^{3\diamond}$, Robert J. Lang$^{3\diamond}$, Nick Robinson$^{3\diamond}$, Brandon Wong$^{4\diamond}$, Marcus Chiam$^{1\star\star}$, Gloria Fang$^{2\star\star}$ and Satinder Singh$^{1\S}$\\
$^1$ Google DeepMind, $^2$ Google, $^3$ Independent, $^4$ Stanford. $^\dagger$ Lead, $^\ddagger$ Tech lead, $^\star$ Core contributor, $^\diamond$ Origami designer, $^{\star\star}$ Partial contributor, $^\S$ Senior lead. Authors in each contribution group are ordered alphabetically.

}
\begin{document}

\begin{abstract}
{\centering

    \includegraphics[width=\textwidth]{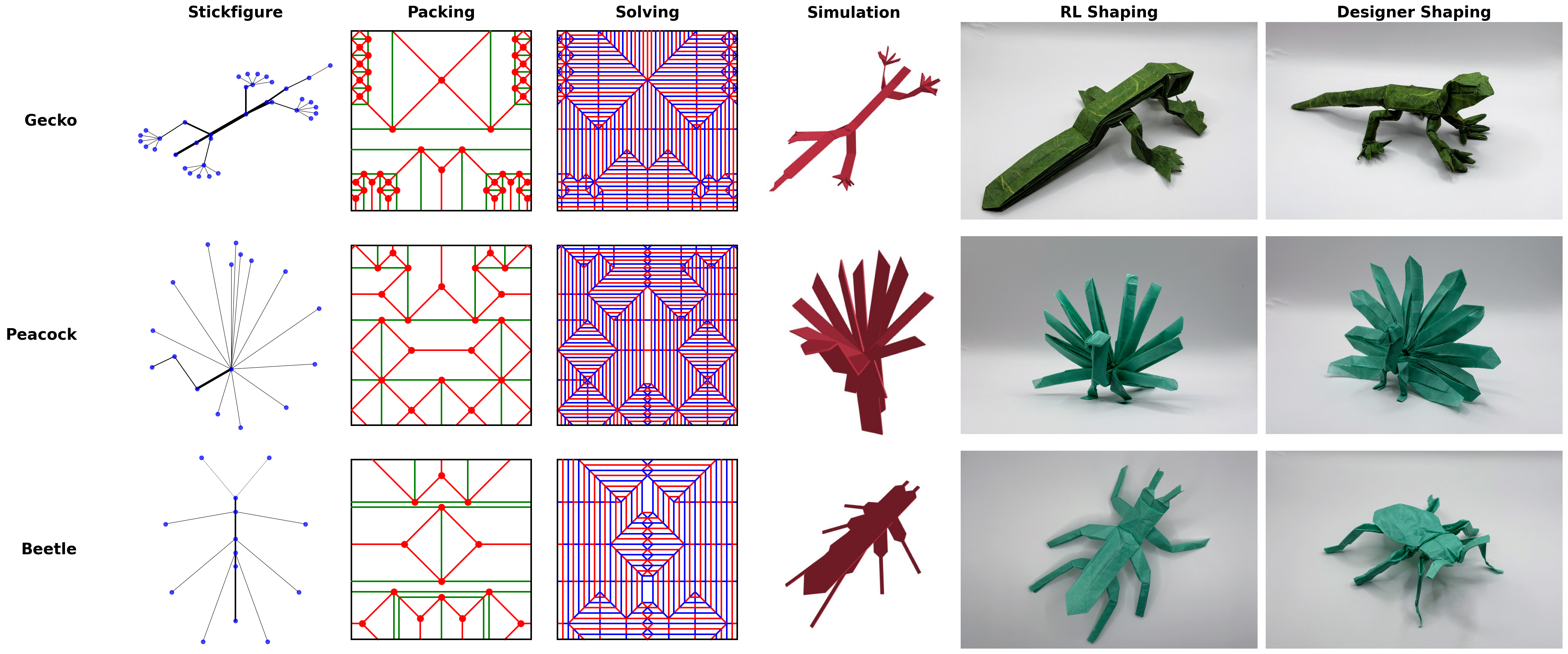}
    \captionof{figure}{Collaborative computational origami with COrigami: progressive design stages bridging natural language to physical art. The workflow transitions from semantic stick figure generation and automated box-pleated crease pattern layout, through simulation and folding, to aesthetic shaping—ultimately establishing a viable structural starting point for physical artist execution by Brandon Wong.}
    \label{fig:teaser}
    \vspace{1em}
    \par} 
While generative AI has achieved remarkable success in solving problems with verifiable solutions, generating physical art that satisfies both strict geometric constraints and subjective visual aesthetics remains a challenge. This paper presents an approach to tackle these difficulties in the domain of computational origami, a mathematically rigid environment that grounds artistic design within the equations of flat foldability. We present COrigami, an end-to-end AI-driven pipeline that assists the design cycle by generating crease patterns from natural language. Our pipeline involves generating a semantic stick figure, computing a base packing, solving for a flat-foldable crease pattern, shaping the flat-folded crease pattern, and refining the generated model using reinforcement learning driven by an autonomous aesthetic evaluation loop. Our system acts as a highly effective collaborative assistant, generating structural starting points that human artists can further expand and shape. By integrating algorithmic optimisation with autonomous aesthetic critique, this work demonstrates how AI systems can satisfy multi-objective physical constraints to enable reliable, mathematically grounded co-creativity.
\end{abstract}

\maketitle


\begin{figure*}[!t]
\includegraphics[width=0.95\linewidth]{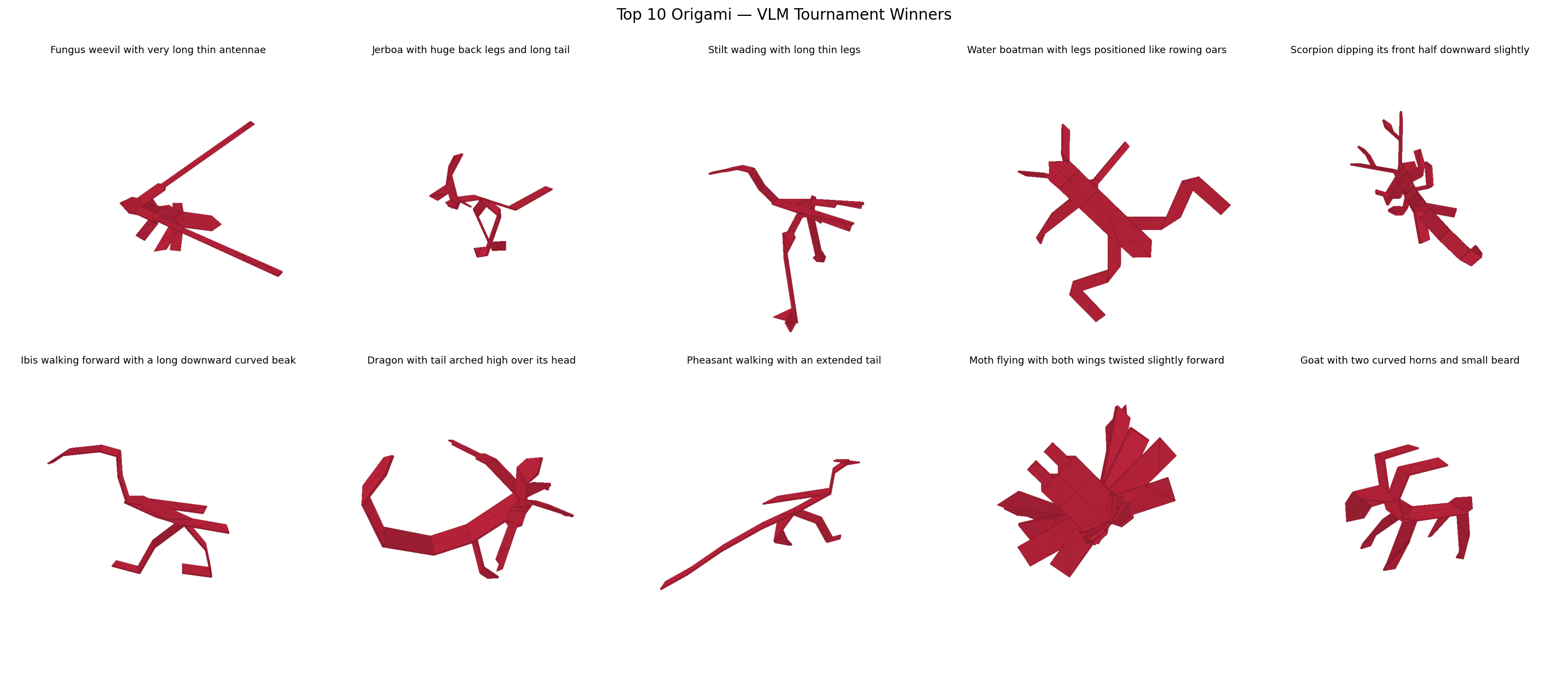}
    \caption{Simulations of origami models designed with COrigami.}
\label{fig:top10}
\end{figure*}

\section{Introduction}
Recent advancements in reinforcement learning applied to large language models have driven rapid progress in domains such as code and mathematics \citep{guo2025deepseek,hubert2025olympiad,Lietal2025,Bhatietal2026}. However, these successes rely heavily on verifiable feedback loops. Translating this reasoning capability to the generation of creative physical art \citep{schmidhuber2010creativity} remains an open challenge as it requires systems to navigate both strict physical viability and subjective human aesthetics.

Origami, the centuries-old art of paper folding, challenges artists to craft expressive representations from a single uncut square paper. Over the past several decades, the complexity of these designs has grown enormously, advancing from simplified forms to highly detailed crustacea, insects, and multi-appendaged creatures. Driven by a modern pursuit of structural realism, models must increasingly mirror the intricate anatomies of real-world subjects. Consequently, the fundamental problem of base design—geometrically configuring the paper to respect a strict target topology—has emerged as the primary bottleneck in the creative workflow \citep{lang1996treemaker}. To address this bottleneck, we investigate a core question: given the powerful semantic reasoning and visual evaluation capabilities of modern LLMs and VLMs, can they be successfully leveraged as collaborative agents to assist and streamline creative origami design?

To understand where AI can contribute, we must first recognise that automated origami design is fundamentally constrained by severe computational hardness. The core geometric challenges of the domain—such as determining whether a crease pattern can fold flat or assigning valid mountain-valley orientations—are provably intractable \citep{bern1996complexity, FlatFolder_OSME2024, Arkinetal2000, Akitayaetal2017, Demaineetal2015, Hull2023}. Historically, generative frameworks attempted to navigate these challenges through continuous spatial optimisation, which produced irrational reference points that were notoriously difficult to execute by hand. To resolve this, modern design has shifted towards discrete box pleating. However, existing interactive editors rely on continuous relaxations that frequently yield non-contiguous gaps requiring intensive manual post-processing. Refer to \cref{tab:comparison} for a high-level comparison of COrigami's features with existing computational origami tools, such as TreeMaker, BP Studio, and Origamizer \citep{demaine2017origamizer}, and to \cref{sec:related_work} for more discussion on related work.

\begin{table*}[t]
  \centering
  \begin{tabular}{l c c c c}
    \toprule
    \textbf{Feature} & \textbf{BP Studio} & \textbf{TreeMaker} & \textbf{Origamizer} & \textbf{COrigami} \\
    \midrule
    Input & Tree & Tree & 3D Mesh & Text \\
    Output & Packing & Base CP & CP & Packing/Base/Shaped CP \\
    Fold Difficulty & N/A & High & Extreme & Low \\
    Automation${^*}$ & Low & Low & Medium & High \\
    \bottomrule
  \end{tabular}
  \caption{Computational origami design tool comparison. ${^*}$Both TreeMaker and BP Studio offer optimization tools for layout design, but they rely on continuous relaxations or manual interactive intervention (e.g., node repositioning, manual symmetry pairing, or resolving non-contiguous gaps) to yield full packing solutions and flat-foldable crease patterns. Only COrigami provides an end-to-end, fully automated pipeline that guarantees contiguous 2D tiling on a discrete orthogonal grid directly from natural language.}
  \label{tab:comparison}
\end{table*}

While recent advancements in multimodal large language models (MLLMs) provide a promising avenue for computational conceptualisation, applying end-to-end generative AI directly to origami design is fundamentally hindered by several compounding challenges. From a generation standpoint, a crease pattern for a visually recognisable model is represented by a highly complex graph containing thousands of shaping creases (see for example Fig \ref{fig:top10}). Because each crease requires tens of tokens to define, the resulting output sequence is exceptionally long. Within this dense topological framework, even minuscule numerical hallucinations or single-token errors cascade into severe flat-foldability violations—a deficit in multi-step spatial reasoning that recent benchmarks, such as OrigamiSpace \citep{xu2025origamispace} and OrigamiBench \citep{agarwal2026origamibench}, have empirically demonstrated in unconstrained frontier models. Indeed, our own preliminary baselines (detailed in \cref{sec:cp_sft_rl}) confirm this architectural limitation; directly fine-tuning frontier models for end-to-end crease pattern generation results in strict flat-foldability plateauing near 60\%.

Furthermore, the domain suffers from a severe scarcity of suitable training data. By consensus within the origami community, crease patterns traditionally serve as abstract structural guidelines rather than exhaustive 3D blueprints, leaving the final aesthetic shaping to the human folder's intuition. Consequently, very few fully fleshed-out, visually recognisable crease patterns exist. As a result, our foundational dataset was limited to approximately 100 visually recognisable traditional origami models developed specifically for this project by collaborating designers. Finally, establishing an autonomous visual reward signal presented a demanding, moving target. Engineering a robust VLM evaluator was uniquely difficult given the initial lack of recognisable baselines and the intricate, multi-perspective geometric details required to capture structural realism in paper folding.

Here we present COrigami, an end-to-end neuro-symbolic pipeline for generating diverse, aesthetic origami designs. Rather than relying on unconstrained generation or continuous spatial relaxations, COrigami systematically produces a crease pattern on a discrete box-pleated grid. Neural models—specifically Gemini and RL—handle the semantic generation and shaping at the beginning and end of the process, while the structural core relies purely on custom algorithmic enhancements to known foldability theorems. This autonomous aesthetic evaluation loop simulates and assesses 3D folds to select models that are simultaneously physically viable and visually compelling. Validated through rigorous ablation studies, COrigami substantially improves upon existing computational origami methods by navigating multi-objective aesthetic and physical constraints. It acts as an effective collaborative assistant that generates reliable structural starting points from which human artists can physically adapt and further shape in their own style (as demonstrated by the designer-folded models in \cref{fig:teaser}). Refer to \cref{tab:comparison} for a high-level comparison of COrigami's features with existing computational origami tools and to \cref{sec:related_work} for related work.
\section{Background}
\label{sec:background}
\begin{figure}[htbp]
    \centering
    \begin{subfigure}[b]{0.42\linewidth}
        \centering
        \includegraphics[width=\linewidth]{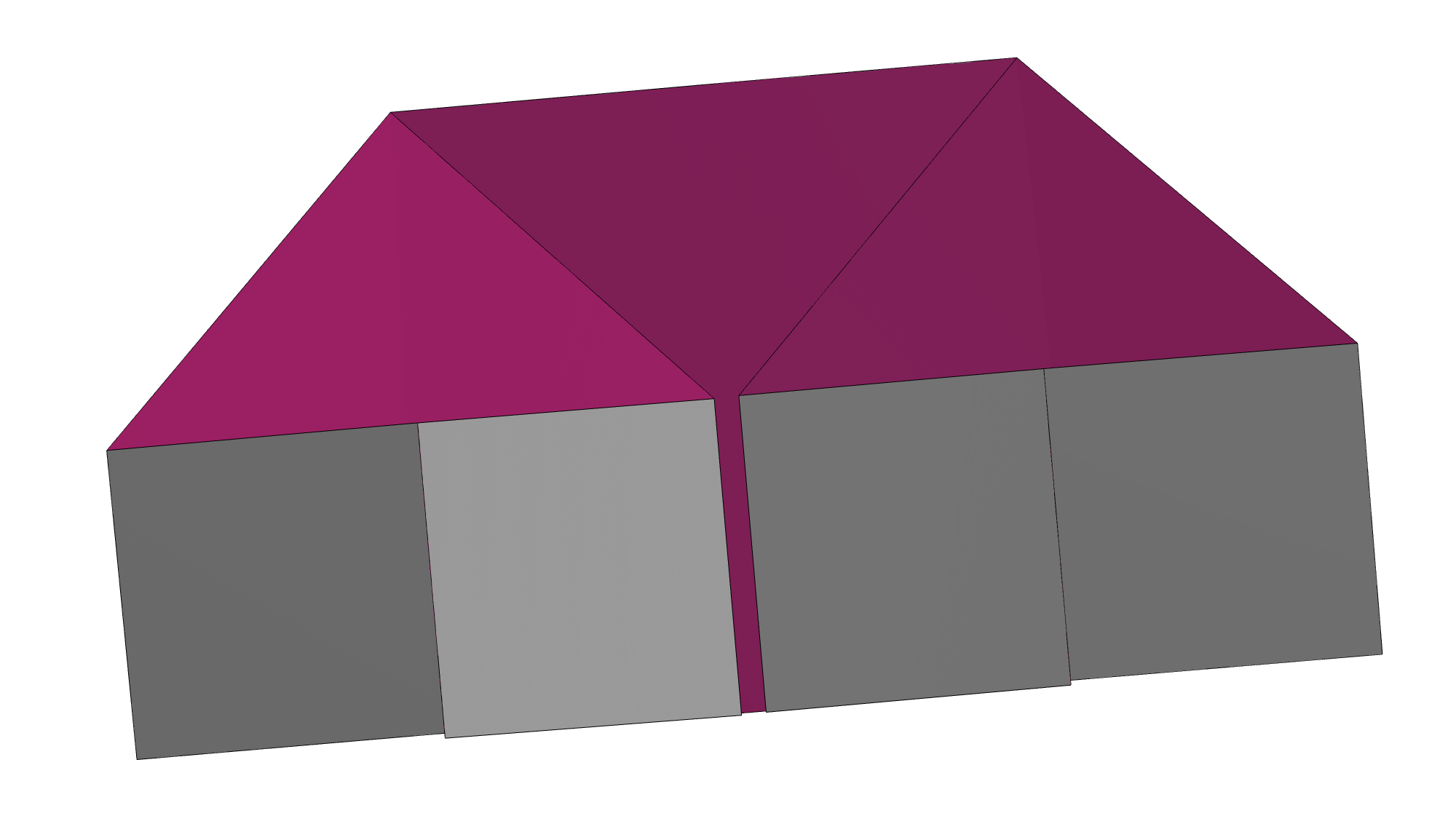}
        \label{fig:sub1}
    \end{subfigure}
    \hfill
    \begin{subfigure}[b]{0.24\linewidth}
        \centering
        \includegraphics[width=\linewidth]{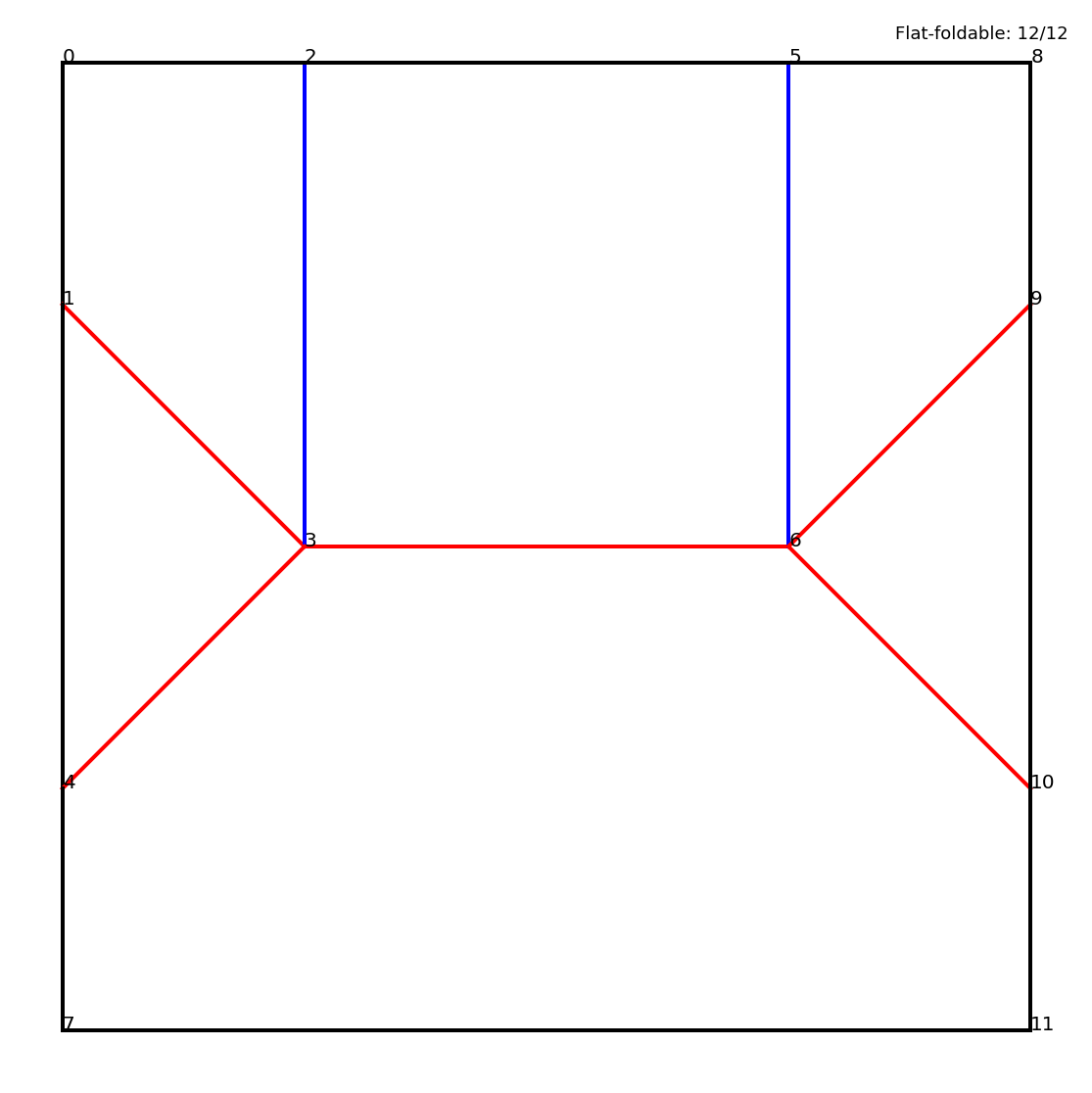}
        \label{fig:sub2}
    \end{subfigure}
    \hfill
    \begin{subfigure}[b]{0.28\linewidth}
        \centering
        \includegraphics[width=\linewidth]{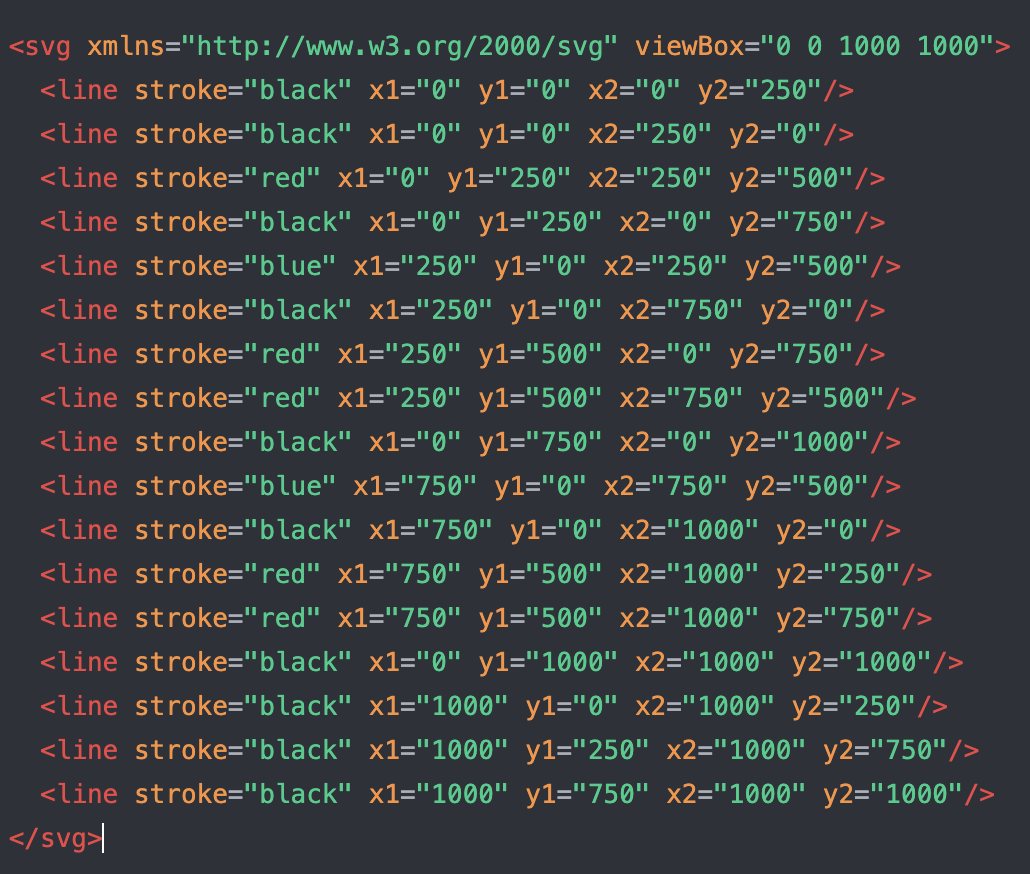}
        \label{fig:sub3}
    \end{subfigure}
    
    \caption{Simulation of an origami house, crease pattern, and the underlying SVG code.}
    \label{fig:cp}
\end{figure}

\textbf{Crease pattern.} \cref{fig:cp} presents a simulation of a traditional origami house (left). Unfolding and flattening the model yields the original flat sheet, revealing a network of outward-pointing (mountain, {\color{red}red}) and inward-pointing (valley, {\color{blue}blue}) folds, as illustrated in the center. This ``crease pattern'' serves as a geometric blueprint for the origami model, dictating the position and orientation of all folds, and can be computationally represented as SVG code (right).

Within our framework, the crease pattern is formalized as a collection of creases and vertices. A crease is defined by its two endpoints, a discrete fold assignment (Mountain (M) or Valley (V)), and a scalar fold percentage in $[0, 1]$. The vertices represent the collection of all crease endpoints and intersections; each vertex explicitly maintains information about its incident creases and the sector angles ($\theta$) between them. When a crease pattern object is instantiated, creases are systematically divided into separate, non-intersecting segments, and any overlapping elements are removed. This explicit data structure allows us to reliably compute flat-foldability conditions at every internal vertex, as described next.

\textbf{Flat Foldability.}
Intuitively, a crease pattern is flat-foldable if the paper can be pressed completely flat into a 2D plane along the crease lines without tearing or self-intersecting. The verification is divided into local and global conditions. Local foldability relies on Kawasaki's and Maekawa's theorems \citep{kasahara1987origami, justin1986mathematics, kawasaki1991relation}, which establish mathematical constraints on the orientations and angles of the crease lines at each vertex. These are supplemented by a recursive crimping algorithm \citep{demaine2007geometric} to ensure valid mountain-valley assignments. Global foldability—verifying the absence of continuous self-intersections—is strictly NP-hard, but can be practically resolved by mapping overlapping convex faces to a finite constraint-satisfaction graph \citep{FlatFolder_OSME2024}. We defer the rigorous mathematical definitions and algorithmic implementations of these physical verifications to \cref{sec:flat_fold}.

\begin{figure*}[t]
    \centering
    \includegraphics[width=\textwidth]{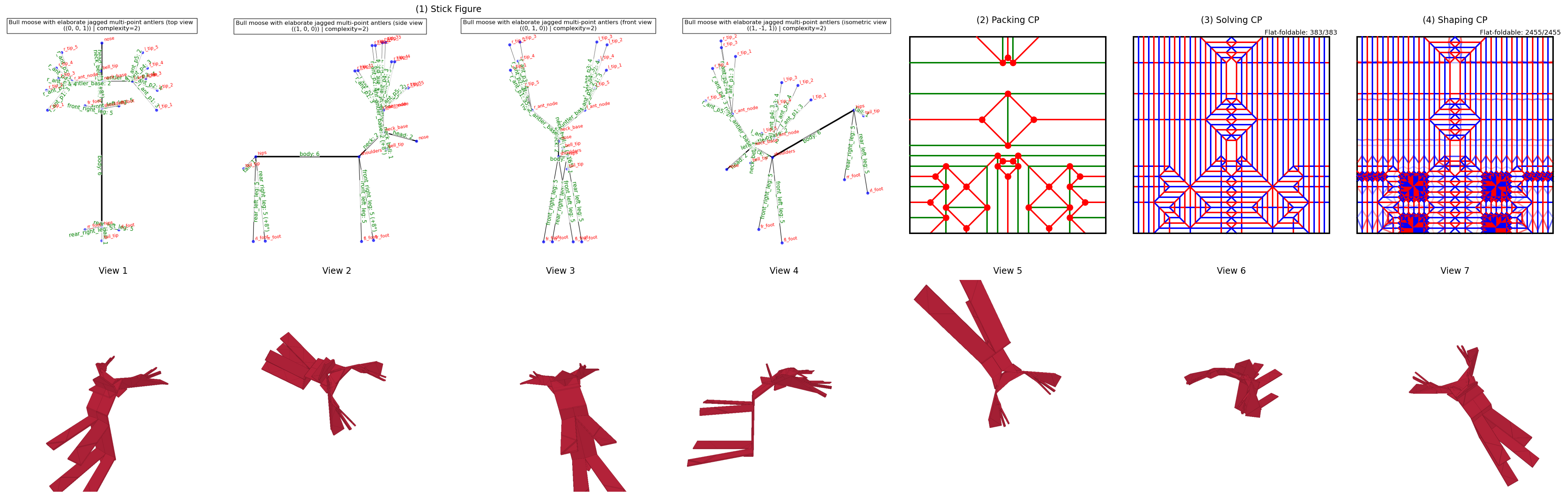}
    \caption{COrigami design process. (0) Gemini is provided the category Cervidae and samples the subject "Bull moose with elaborate jagged multi-point antlers." (1) The system generates a semantic stick figure, shown from four viewing angles. (2) The initial packing pattern is generated, indicating hinge creases (green) and ridge creases (red). (3) The solved crease pattern assigns mountain (red) and valley (blue) assignments to the pleats, ridges, and hinges. (4) The final shaped crease pattern is produced. Finally, the model is folded and presented in seven perspectives.}
    \label{fig:box-pleating}
\end{figure*}

\textbf{Origami design.}
Traditional origami is historically defined by the strict constraint of folding a single, uncut sheet of square paper into a recognizable representation of a subject. For centuries, design was guided by intuitive, trial-and-error folding sequences. Over time, these practices were formalized into standard structural starting points known as traditional bases. Each base is a pre-configured crease layout that systematically distributes the paper's area into a fixed number of terminal flaps, which correspond directly to the subject's primary appendages. To overcome the rigid limits of classical origami bases, 20th-century designers developed systematic manual techniques—such as splitting, grafting, and tiling—to dynamically modify, extend, and combine underlying crease patterns \citep{lang2012origami}.

As modern designers pursued highly detailed, multi-appendaged anatomies, these classical bases reached their limits, pushing the field to explore computational origami. Early algorithmic approaches formalized flap allocation through continuous optimization frameworks like circle packing \citep{lang1996treemaker}. However, these continuous layouts yield irrational folding angles that are extremely difficult to execute by hand. Furthermore, simply constraining a design to a finite set of angles does not fully resolve the issue; continuous spatial routing can still trigger a geometric trap known as 'infinite bouncing' \citep{Demaine1998, Demaine1999, lang2012origami},  where crease propagation never terminates.

To resolve these barriers, modern computational design transitioned to discrete "box pleating", which restricts all axis-parallel creases and hinges strictly to an orthogonal integer grid and $45^{\circ}$ diagonal creases. Crucially, this grid framework provides two fundamental advantages: the angle constraint guarantees rational, reproducible folding angles; but more importantly from a designer's perspective, it mathematically ensures finiteness of the constructed creases. Discretising the design space maps continuous geometric packing to discrete combinatorial state-space searches, guaranteeing both physical reproducibility for human folders and computational tractability for automated pipelines like COrigami. For a detailed discussion on related work for origami design, we refer the reader to \cref{sec:related_work}.

\section{Methods}
As illustrated in \cref{fig:box-pleating}, the COrigami design pipeline is an end-to-end neuro-symbolic system combining AI-driven design with algorithmic tools for box-pleating crease pattern construction and shaping. Crucially, while expert human designers manually pack and solve box-pleated bases, no software previously existed to fully automate this contiguous tiling and flat-foldability process. Our pipeline introduces the first fully automated algorithmic implementation and evaluation of these techniques.

The pipeline operates as follows: First, it leverages a Gemini-based \citep{GeminiTeam2025Gemini3} AI workflow to convert a user prompt (e.g., a \textit{bull moose with elaborate jagged multi-point antlers} in \cref{fig:box-pleating}) into a semantic stick figure. This defines both the topology of the box-pleating design and provides high-level direction for subsequent shaping. Next, we introduce custom-built algorithms to convert the semantic stick figure into a 2D crease pattern: our backtracking solver converts this stick figure into a discrete 2D rectangle \textit{packing}, and our hinge algorithm \textit{solves} it to produce a guaranteed flat-foldable base crease pattern. We then proceed to \textit{shaping} the model in two stages. First, a novel "tree shaping" algorithm uses angle information from the stick figure to apply simple folds, pushing the flat base into a 3D posture that matches the rigid stick figure skeleton. Second, Gemini is reintroduced with a Reinforcement Learning (RL) framework to orchestrate tools for additional shaping techniques to improve this blocky skeleton, refining the model to resemble the actual target subject. Finally, a custom geometric folding simulator renders the shaped crease pattern from multiple perspectives; these perspectives are used to provide VLM feedback as the reward signal for RL.

\subsection{Stick figure generation.} 
\label{subsec:sf}
The first step in the pipeline is to create a semantic stick figure, which provides a guideline for the final origami design. A semantic stick figure operates similarly to a traditional topological tree, but provides textual descriptions of the design components and orientations in 3D space. The stick figures are created by Gemini using a constrained prompt workflow, which enforces specific topological properties, such as symmetry and absence of graph cycles. Each edge (or "stick") is explicitly parametrised by a unique label and spatial properties: scalar length, azimuth angle, and elevation angle. Within the framework of box-pleating, leaf nodes map directly to flaps (terminal appendages connected at a single vertex) and internal edges map to rivers (internal structural connectors). This tree data structure allows us to compute important properties for downstream tasks, such as stick adjacency and grid-size approximations. 

\textbf{VLM-Driven Refinement.} We use Gemini as a Vision-Language Model (VLM) to verify that the stick figure aligns with the intended target. The VLM is provided with the JSON description along with four 3D renderings of the stick figure from different viewpoints: Top, Side, Front, and Isometric.
It evaluates the topology against four discrete criteria: topological accuracy (validating the vertex/edge count against the semantic target), proportional feasibility, semantic recognisability, and structural complexity. 
If the design score is too low, the LLM is asked to refine the stick figure design. 
To guide this refinement, we explicitly instruct the model to adjust stick lengths for accurate proportions, correct joint angles to resolve geometric overlaps, enforce mirrored angles for symmetric pairs (e.g., left and right limbs), and add or remove nodes to better align with the target topology.

\begin{figure*}[t]
    \centering
    \includegraphics[width=\textwidth]{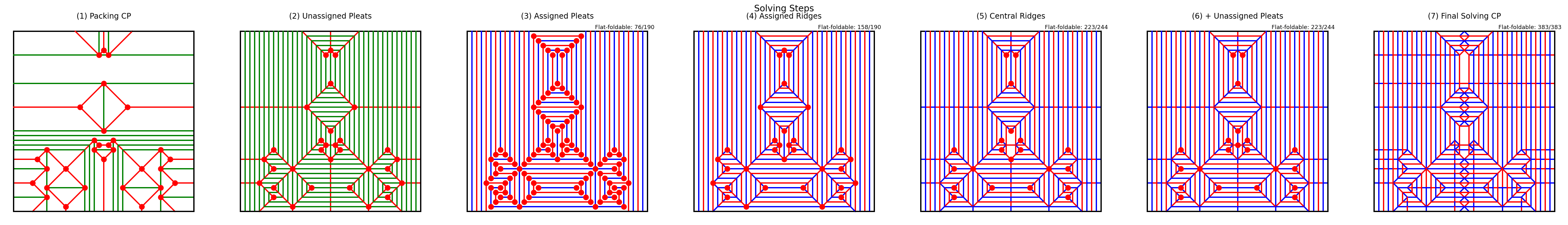}
    \caption{\textbf{Deterministic solving steps.} The process begins with an initial packing crease pattern (1). Next, unassigned pleats are generated (2) and subsequently assigned (3). General ridges are then assigned (4), followed by the central ridges (5). Specific pleats are removed to be reassigned later (6). This is a final solution of the hinge solver (7).}
    \label{fig:solving}
\end{figure*}

\subsection{Packing}
In this stage, the semantic stick figure is mapped onto a square paper by solving a discrete rectangle-packing and tiling problem on a square integer grid. The grid size is heuristically estimated from the stick figure, with the solver subsequently evaluating multiple candidate grid sizes (see \cref{app:grid_size_heuristic} for further details). Leaf nodes of the topological tree are instantiated as rectangles whose dimensions are derived from the corresponding stick lengths, representing anatomical flaps, while internal edges become paths of proportional width representing structural rivers. Rivers partition the grid into bounded regions called \emph{pockets}, each containing a subset of flaps that must be packed together. This mapping is strictly governed by graph adjacency: components sharing a joint in the tree must belong to the same pocket in the planar layout. The packing solver operates as an iterative backtracking search over river placements and flap positions. Rivers are placed sequentially according to a traversal plan---an ordering of tree edges that respects topological dependencies. The first river is placed from an exhaustive enumeration of candidates that span the full grid width or height (straight rivers) or include a single $90^{\circ}$ turn (L-shaped rivers). Each subsequent river is placed using a wall-following algorithm that traces the contour of previously placed elements, producing free-form paths that snake around existing obstacles. After each river is placed, its newly enclosed pockets are immediately packed with their constituent flaps. Candidate flap positions are produced either analytically---by sliding the rectangle along the faces of adjacency-constrained neighbours---or via brute-force grid enumeration, and are filtered through symmetry, overlap, and area-feasibility checks before being ranked by a scoring heuristic.

\textbf{Tiling (via Flap Expansion)}.
Once a structurally valid preliminary layout is established, the algorithm must eliminate all residual gaps to achieve a perfect tiling---a mathematical prerequisite for generating a physically viable base crease pattern. The system identifies all unoccupied cells and, for each, computes feasible geometric expansions of adjacent flaps that will cover that cell. A backtracking search over these candidate expansions finds a consistent, non-conflicting set of flap enlargements that fills every gap.

Ultimately, this process yields a complete packing layout where diagonal ridges---internal creases representing the straight skeleton of each region---structurally partition the paper into discrete regions. This layout also defines the initial geometry of hinges---fold lines where adjacent regions meet---for the crease pattern, achieving an optimal use of the paper by minimizing the required grid size for a given stick figure.

\subsection{Solving}
As illustrated in \cref{fig:solving}, the algorithm starts from a packing layout, with unassigned ridge and hinge creases (step 1 in \cref{fig:solving}), and produces a flat-foldable crease pattern in a sequence of steps. The ridges structurally partition the paper into discrete regions. The algorithm uses this geometric partition, and constructs pleats within each region such that they share a uniform orthogonal orientation—either strictly horizontal or vertical. This axial alignment is systematically resolved using a five-step geometric filtering protocol (detailed in \cref{subsub:pleats_con}).

With all the creases constructed (step 2 in \cref{fig:solving}), the method transitions to the Mountain/Valley (M/V) assignment of the pleats and ridges. Pleat segments grouped into connected paths are assigned an identical fold orientation, while adjacent parallel components are strictly interleaved (alternating Mountain and Valley, as can be seen in step 3 in \cref{fig:solving}). Subsequently, ridge orientations are deterministically anchored at highly constrained junctions (e.g., Y-shape vertices and paper boundaries) and propagated through the network, governed by intersection rules that enforce alternating parity at each vertex. Upon the conclusion of this stage, the majority of the vertices within the crease pattern already inherently satisfy local flat-foldability conditions. This deterministic phase thereby establishes a rigid mathematical foundation, successfully isolating the remaining exponential complexity of the problem to the unassigned hinge creases (step 4 in \cref{fig:solving}).

\textbf{Combinatorial Hinge Assignment.} The last step involves a combinatorial assignment problem: A subset of the unassigned hinge creases has to be assigned in order to produce a flat foldable crease pattern. The system executes a priority-driven, greedy state-space search. At each step, the algorithm generates new candidate states by assigning each of the hinges with an orientation and appending them to a queue. These assignments take one of two forms: \textbf{interleaved} (a standard M-V-M-V zigzag fold) or \textbf{symmetric} (an M-V-V-M assignment where central segments share a fold). The queued states are strictly prioritised by a global score, defined as the sum of individual vertex scores that quantify how close the pattern is to complete local flat-foldability; any scoring ties are resolved by evaluating hinge length and symmetry. A critical innovation during this state generation is the flexible handling of pleat reassignment. When a symmetric hinge is assigned, it effectively ``traps'' a pleat, moving it from the fixed pattern back into the \texttt{unassigned\_pleats} pool. This deferred decision-making allows the solver to dynamically flip the pleat's orientation downstream, ensuring it can satisfy local constraints at any vertex the pleat traverses.

To circumvent the inherent complexity of this combinatorial assignment, the spatial geometry is decomposed into disjoint, hierarchical partitions. Each partition represents a connected component constructed via traversal: hinges are grouped together if they are connected through shared vertices, and each connected component corresponds to a specific joint in the uniaxial tree. Crucially, each partition stores its own isolated set of vertices, enabling the solver to calculate local flat-foldability scores independently of the global pattern.

Once constructed, these partitions are sorted in descending order of vertex count, forming a sequential solving queue that tackles the most highly constrained clusters first. To further maintain efficiency, aggressive algorithmic pruning immediately discards candidate states that trigger strict failure conditions, such as greedy score degradation $(S_{current} < S_{parent})$, grid-based structural misalignments, or mathematically unsolvable vertices where local Maekawa theorems cannot be satisfied. Ultimately, this combination of localised partitioning, heuristic guidance, and robust pruning ensures efficient convergence to a globally valid crease pattern.

\subsection{Shaping}
The solving phase produces what is known in box-pleated design as a collapsed base. Because all the structural flaps are compactly folded on top of one another to satisfy the 2D flat-foldability constraints, this base physically resembles a dense, flat-folded strip (or tube) of paper. Crucially, however, this strip internally preserves the exact branching topology of the original stick figure and serves as a rough structural skeleton. To produce the final model, we must add a series of shaping creases to this base crease pattern. This process is divided into two distinct stages: an algorithmic stage that uses simple folding to recreate the 3D geometry of the stick figure, and a subsequent RL-driven stage orchestrating additional shaping techniques. The ultimate objective of this second phase is to orchestrate complex folds that match the aesthetic and semantic features of the actual target subject (e.g., tapering a thick flap into an insect's delicate leg) using VLM feedack.
\subsection{Shaping techniques}
\textbf{The Simple Fold.} The simple fold is one of the most basic origami folds. Given a \textit{cut line} defined by two points $p_1$, $p_2$ the simple fold folds the paper over the cut line as either a mountain or valley fold. Despite its simplicity, this basic fold is highly expressive, allowing us to do complex 3D shaping. We implement the simple fold by checking the intersection of each flat-folded \textit{face} $f_i$ of the folded base crease pattern with the cut line, handling numerical error near vertices carefully. Each face intersection is recorded as two face-boundary intersection points $(f_{i,1}, f_{i,2}) \in \mathbb{R}^2 \times \mathbb{R}^2$. Each pair forms a shaping crease segment on the crease pattern with the intersection points as endpoints and assignment determined by the Mountain/Valley fold type of the simple fold and the orientation of the folded face. Additional support is added for selectively shaping subsets of the crease pattern corresponding to distinct flaps/rivers. This allows us to apply distinct simple folds to flaps/rivers that flat-fold on top of each other in the base crease pattern.

\textbf{Narrowing and the Clip Pattern Algorithm.} While the simple fold effectively handles rigid 3D directional changes, anatomical realism often requires more complex shaping such as narrowing (e.g., tapering an insect limb). We introduce the \textit{clip pattern algorithm}, a framework for applying complex 2D crease templates---such as narrowing---across multi-layered folded flaps. 

The clip pattern algorithm propagates a local 2D reference frame across layers of a flat-folded flap. The system selects a reference face and initializes a virtual drawing machine parametrized by local homogeneous transformations and a crease template. For every other face of the flap, we compute an affine transformation by traversing the topological folding path from the reference face, ensuring the template is accurately projected onto the target layer within the unfolded 2D base. Crucially, the algorithm automatically detects Z-axis flips along the folding path, dynamically swapping Mountain/Valley assignments of every crease in the template to maintain physical consistency. All generated lines are geometrically clipped to the hull of their respective faces, yielding a consistent set of shaping creases. For a narrowing template, the flap's width becomes smaller without changing its direction. The algorithm works for rivers too.
\subsection{Orchestrating Shaping}
\textbf{Algorithmic Orchestration.} For orchestrating the simple fold tool, we developed the \textit{tree shaping} algorithm  which converts a generated stick figure into a series of simple folds via a breadth-first-search traversal from the root stick. For each un-shaped section of paper (flap/river) on the BFS frontier, the algorithm computes the necessary simple fold required to align the paper with the corresponding target stick's orientation in the stick-figure. The corresponding rotation of the parent stick to the target stick is subject to the strict geometric constraint that the rotation axis must be physically realizable as a simple fold cut line on the parent's paper plane. Once the orientation of the root flap/river in 3D is chosen, this completely determines the necessary simple folds for all other flaps/rivers. We provide a detailed description in \cref{sec:shaping_supp}.

\textbf{Reinforcement learning.}
While our heuristic algorithm successfully maps geometric orientations to simple folds, it strictly aims to perfectly reconstruct the original stick figure. This faithful translation cannot guarantee an aesthetically pleasing 3D model, and the initial LLM-generated stick figures may themselves contain structural or proportional flaws. To overcome this limitation and bridge the gap between the rough topological skeleton and the actual real-world subject, we leverage Gemini (specifically Gemini 2.5 Flash Lite \citep{comanici2025gemini}) to generate the parameterization for the shaping tool orchestration. To simplify the optimization problem, we formulate this as a single-step execution pipeline: provided with the stick figure specifications, in-context examples, the available shaping tool declarations, and a comprehensive instruction prompt, the model is tasked with outputting the exact tool-use parameters for all flaps simultaneously in one step. These parameters are then executed in our simulator to produce the final model.

Because the provided context is highly detailed, standard prompting allows the model to easily map the stick figure parameters directly to the shaping tools—a relatively straightforward task for Gemini that effectively replicates the baseline performance of the heuristic algorithm. To further promote shaping tool orchestration and improve over these prompting baselines, we apply a reinforcement learning (RL) framework to this generative pipeline and finetune Gemini 2.5 Flash Lite for the orchestration task. During this RL phase, the agent navigates a broader and more expressive action space, empowering it to selectively narrow specific sections of the paper and apply additional simple folds to fine-tune the model's anatomy. The learning process is steered by a structured reward formulation that balances aesthetics with strict physical constraints. Specifically, the agent receives a penalty reward of $r=-1$ if the generated trajectory is invalid (e.g., if no tool is called, the resulting shaped model violates flat-foldability or triggers a simulation error). Otherwise, the reward is determined by the VLM-derived aesthetic feedback (Single Model Evaluation Mode, which will be discussed in \cref{sec:vlm_feedback}), coupled with an intrinsic reward for action diversity ($r_i = \min(\frac {n}{10}, 1) * 0.6$, where $n$ is the number of successful tool calls) to promote comprehensive exploration. To ensure physical realisability, the policy is trained exclusively on structurally valid outcomes that adhere to formal crease syntax, maintain flat-foldability, and yield minimal simulation error.
\subsection{Folding}
Although several origami simulation tools exist, they generally focus either on 2D foldability analysis \citep{mitani2005oripa, FlatFolder_OSME2024} or on dynamic, physics-based simulation \citep{ghassaei2018fast}. When integrated into an automated pipeline, we found that the latter approach tends to accumulate relatively high strain errors. To address this limitation, we developed a novel, purely geometric folding simulator. We refer the reader to \cref{sec:folding_supp} for an empirical comparison of our approach against existing physical simulators.

The folding phase deterministically constructs the 3D geometry of the folded model from the 2D crease pattern. The system first parses the pattern into lists of vertices, edges, and faces, sanitising the topology by eliminating dangling elements and validating face connectivity. To compute the spatial configuration, the algorithm constructs a face-adjacency graph, where two faces are adjacent if they share an edge, and executes a breadth-first traversal rooted at an arbitrary face. The 3D coordinates are deterministically resolved by computing global $4 \times 4$ affine transformations for each face based on its specified fold angle. Because a single vertex typically belongs to multiple faces, its final 3D coordinate is resolved by averaging its transformed positions across all faces containing it to mitigate floating-point inaccuracies. Finally, to evaluate geometric consistency and detect foldability conflicts, we compute the mean axial strain, which measures the relative change in edge lengths between the original 2D and folded 3D states.

\subsection{VLM feedback}
\label{sec:vlm_feedback}
To bridge the gap between rigorous geometric validity and subjective artistic quality, we utilize a Vision-Language Model (VLM), specifically Gemini 3 Flash \citep{GeminiTeam2025Gemini3} with temperature of $0$ and no majority voting (single call), as an automated aesthetic and semantic judge. This design shares a notable architectural parallel with recent work by \citet{banarse2026evolution}, who deployed a multimodal model within a generative 3D pipeline to act as an automated curator capable of mapping open-ended semantic targets directly to aesthetic selections. While a tree similarity score (defined in \cref{subsec:tree_sim}) effectively measures how closely the folded model aligns with the target stick-figure skeleton, it is fundamentally limited to rigid geometric reconstruction. A mathematically faithful translation of a stick figure does not guarantee an aesthetically pleasing 3D model, especially if the initial generated skeleton contains proportional flaws. Furthermore, stick figures cannot represent critical dimensional features like appendage width. The VLM overcomes these limitations by evaluating the final model from seven different views, allowing the downstream reinforcement learning phase to deviate from the rigid skeleton and apply advanced narrowing techniques to achieve true structural realism. Our VLM evaluation pipeline operates in two distinct modes:

\textbf{Single Model Evaluation Mode.} In this mode, the VLM is provided with the textual prompt and the seven rendered images of a single candidate model. The model is tasked with performing multi-angle spatial reasoning to assess how well the folded geometry represents the semantic features of the target object (e.g., the "jagged points" of a moose's antlers or the "tapered tail" of a cat). The VLM generates a detailed chain-of-thought analysis before providing a final correspondence score.

\textbf{Comparison Judge Mode.} We also implement a comparative evaluation framework. In this mode, the VLM is presented with two images—such as different views of the same model or alternative shaping attempts of the same subject. The VLM performs a side-by-side structural comparison to determine which model exhibits superior fidelity to the text description. To mitigate proximity bias, we evaluate the models twice by swapping their presentation order. Refer to \cref{apx:vlm_feedback_prompt} for the prompt templates used in both modes.

\textbf{Scoring and Integration.} In both modes, the VLM is prompted to output a discrete numerical score ranging from 0 to 10 after the reasoning process. Empirically, we found that single-model evaluation efficiently filters out erroneous or visually flawed models, while the comparison mode is more effective at identifying the highest-quality outputs. To integrate this feedback into our computational pipeline, we normalize these values to a continuous range of $[0, 1]$. This strategy provides a robust and consistent reward signal that serves a dual purpose: (1) it acts as the primary selection criterion within the pipeline to identify the most promising origami models using shaping heuristics, and (2) it functions as the reward signal for the Reinforcement Learning shaping process, guiding the discovery of intricate 3D folds that maximize correspondence, realism, and aesthetic recognizability. Because the VLM evaluates rendered 3D meshes rather than physical paper models, its aesthetic judgment does not account for the structural bulk or thickness limitations of physical paper. Consequently, models that receive high scores in simulation may still require significant physical post-processing and manual layer thinning by the human artist to achieve the same visual clarity in physical form.
\section{Experiments}
Before evaluating the components of our proposed neuro-symbolic pipeline, we first established an empirical baseline to test the limits of unconstrained generative architectures. As detailed in \cref{sec:cp_sft_rl}, directly fine-tuning a language model to output raw crease patterns in SVG space results in a hard performance ceiling; while structural syntax validity improves during training, strict mathematical flat-foldability plateaus near 60\%. Having confirmed that end-to-end generation is insufficient, the following experiments evaluate our decoupled, discrete box-pleating approach.

\subsection{VLM Evaluator Benchmarking}
\begin{table}[t]
\centering
\setlength{\tabcolsep}{3pt} 
\caption{Unified benchmark results across model architectures, sampling budgets ($N$), and prompt variants on VLM origami evaluation dataset. We report classification Accuracy, Average Precision (AP), and $F_1$ Score at optimal score thresholds. Part A sweeps models and samples ($N$) under Rubrics baseline prompt. Part B sweeps prompt templates using Flash at $T=0.0, N=1$. Winner rows are typeset in bold.}
\label{tab:vlm_unified_master}
\begin{tabular}{lcccccc}
\toprule
\textbf{Model} & \textbf{T} & \textbf{N} & \textbf{Prompt} & \textbf{Acc} & \textbf{AP} & \textbf{F$_{1}$} \\
\midrule
\multicolumn{7}{l}{\textbf{Part A: Model \& Sampling Sweeps}} \\
\addlinespace
Flash & 0.0 & 1 & Rubrics & 0.715 & 0.565 & 0.640 \\
& 1.0 & 1 & Rubrics & 0.749 & 0.602 & 0.688 \\
& \textbf{1.0} & \textbf{4} & \textbf{Rubrics} & \textbf{0.766} & \textbf{0.665} & \textbf{0.689} \\
& 1.0 & 16 & Rubrics & 0.741 & 0.652 & 0.669 \\
\addlinespace
Pro & 0.0 & 1 & Rubrics & 0.640 & 0.429 & 0.632 \\
& 1.0 & 1 & Rubrics & 0.644 & 0.418 & 0.635 \\
& 1.0 & 4 & Rubrics & 0.632 & 0.405 & 0.613 \\
& 1.0 & 16 & Rubrics & 0.636 & 0.405 & 0.626 \\
\midrule
\multicolumn{7}{l}{\textbf{Part B: VLM Prompt Sensitivity Study}} \\
\addlinespace
\textbf{Flash} & \textbf{0.0} & \textbf{1} & \textbf{Rubrics} & \textbf{0.715} & \textbf{0.565} & \textbf{0.640} \\
& & & Rubrics, V0 & 0.631 & 0.465 & 0.631 \\
& & & Score & 0.632 & 0.470 & 0.632 \\
& & & Binary & 0.590 & 0.455 & 0.611 \\
\bottomrule
\multicolumn{7}{l}{\textbf{Part C: VLM Tournament}} \\
\addlinespace
{Flash} & {0.0} & {1} & {Baseline} & {0.715} & {0.565} & {0.640} \\
& {0.0}& 1& View & 0.736 & 0.561 & 0.645 \\
& {0.0}& 1& {Double} & {0.803} & 0.649 & {0.713} \\
& {1.0}& {4}& View & 0.728 & 0.615 & 0.658 \\
& {1.0}& {4}& \textbf{Double} & \textbf{0.811} & \textbf{0.651} & \textbf{0.74} \\
\bottomrule
\end{tabular}
\end{table} 
Before deploying the Vision-Language Model as an autonomous aesthetic judge and RL reward signal, we rigorously benchmarked its evaluation capabilities. To determine the optimal configuration, we evaluated multiple model architectures, sampling budgets ($N$), and prompt variants on a curated VLM origami evaluation dataset (with 87 positive examples and 152 negative examples). As detailed in \cref{tab:vlm_unified_master}, our ablation study reveals several key insights. First, as shown in Part A, the Gemini Flash architecture surprisingly outperforms the Pro model on this specific spatial reasoning and structural evaluation task. Furthermore, while a greedy decoding strategy ($T=0.0$) provides a strong baseline, introducing temperature scaling ($T=1.0$) combined with a best-of-$N$ sampling budget ($N=4$) yields the optimal performance, achieving a classification accuracy of 0.766 and an $F_1$ score of 0.689. Crucially, Part B demonstrates the strong impact of prompt engineering on model performance. The highly structured "Rubrics" prompt—which forces the model to explicitly verify appendage counts, topology, proportionality, and differentiation before scoring (see Section 3.6)—vastly outperforms Rubrics, V0. To get from V0 to the final version, we manually inspected many samples with origami designers and refined the prompt to reflect their preferences. In addition, we can see that the Rubrics prompt outperforms simpler zero-shot prompts (such as basic scoring or binary pos/neg evaluations). Lastly, Part C highlights the effectiveness of side-by-side VLM comparisons using tournament-style evaluations. The 'View' method—which first conducts a tournament to select the optimal viewing angle before applying the Rubrics prompt—yields marginal improvements over the baseline. However, the 'Double' tournament approach provides a substantial performance leap. This method introduces a second stage, pitting the best views of different models against one another in a direct comparison, with binary classification derived from an optimal ranking threshold. This dual-tournament strategy achieves a striking 0.811 classification accuracy, 0.651 average precision and an $F_1$ score of 0.74. Given its superior discriminative power, we deployed this double tournament in practice to reliably curate the top-performing final designs. These benchmark results directly informed the VLM configuration for our generative pipeline and RL orchestration loops. Since performance gains at higher parameters were negligible, we defaulted to $T=0.0$ and $N=1$ for a computationally efficient, deterministic, and reproducible setup.

\subsection{Algorithmic Generation of Origami Models}
The proposed generative framework operates in two distinct stages. In the first phase, the system executes the complete end-to-end pipeline—initiating with the generation of semantic stick figures and concluding with initial 3D shaping driven entirely by algorithmic heuristics. The primary objective of this initial stage is to efficiently explore the design space and distil a robust, curated dataset of candidate models that will be further refined using Reinforcement Learning (RL) in the second stage.

\begin{figure}[h]
    \centering
    \includegraphics[width=\columnwidth]{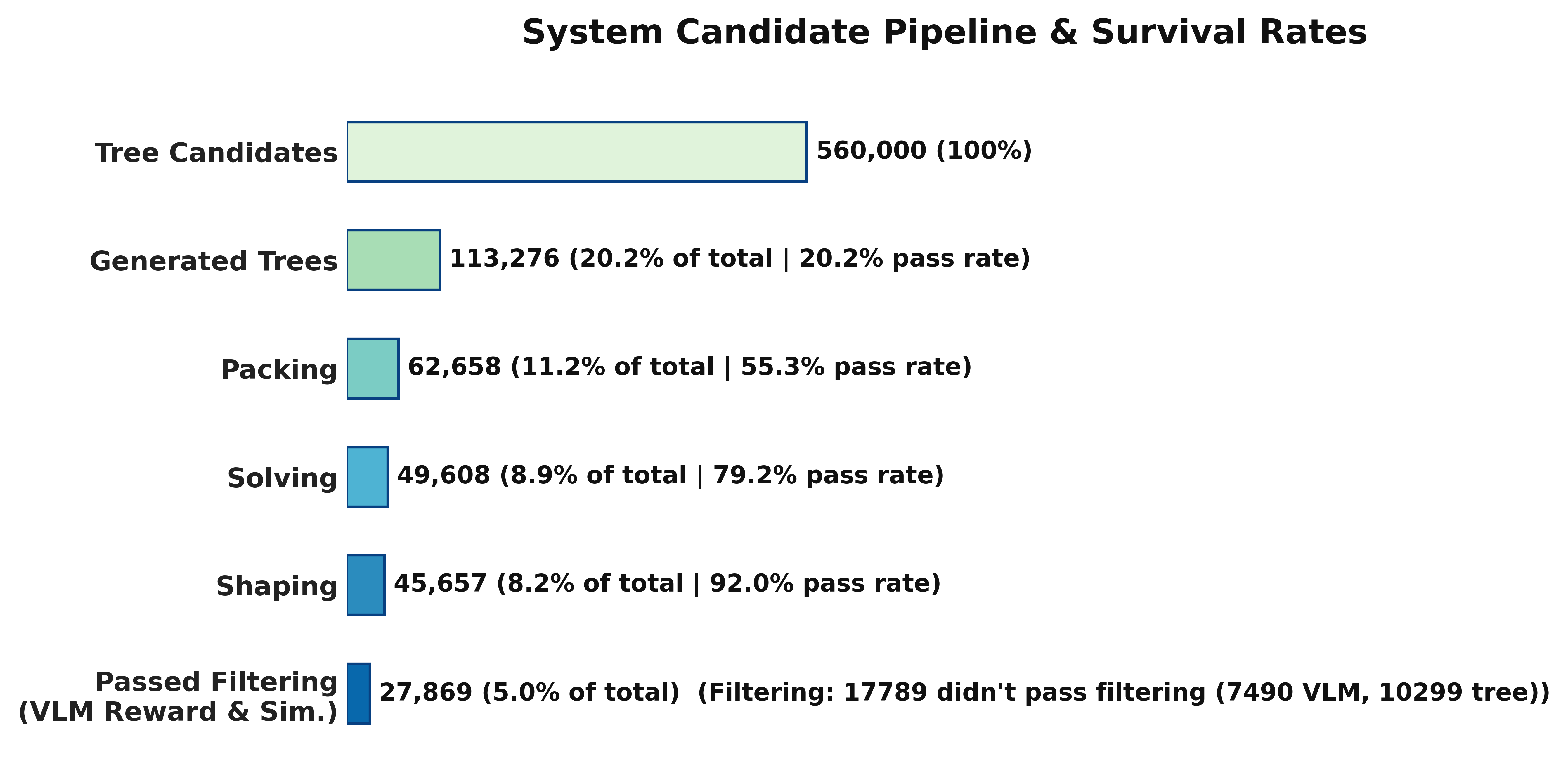}
    \caption{Success rates throughout our pipeline.}
    \label{fig:tq_summary}
\end{figure}

To guarantee high-quality outputs, this first stage actively explores a large number of packing and solving solutions for each generated stick figure. Packing variations are achieved by sweeping across different grid resolutions, while diverse crease patterns are produced by varying key solver hyperparameters and exploring multiple valid hinge assignments per layout. The initial grid resolution is a lower bound derived from circle-packing heuristics that estimates required paper area from stick lengths and types, clamped below by the tree diameter (see \cref{app:grid_size_heuristic} for details). The sweep then increments this base resolution in unit steps, re-attempting packing at each successive grid size until a valid tiling is found or a maximum bound is reached. The optimal candidate for each stick figure is rigorously evaluated and selected using the VLM in the single model evaluation mode, with a topological tree-similarity score employed as a definitive tie-breaker. Ultimately, the top-performing models are filtered to have VLM score larger than $0.6$ and tree similarity score (see \cref{subsec:tree_sim} for details) larger than $0.9$ to construct the curated dataset required for the downstream RL.

\begin{figure}[h]
    \centering
    \includegraphics[width=\columnwidth]{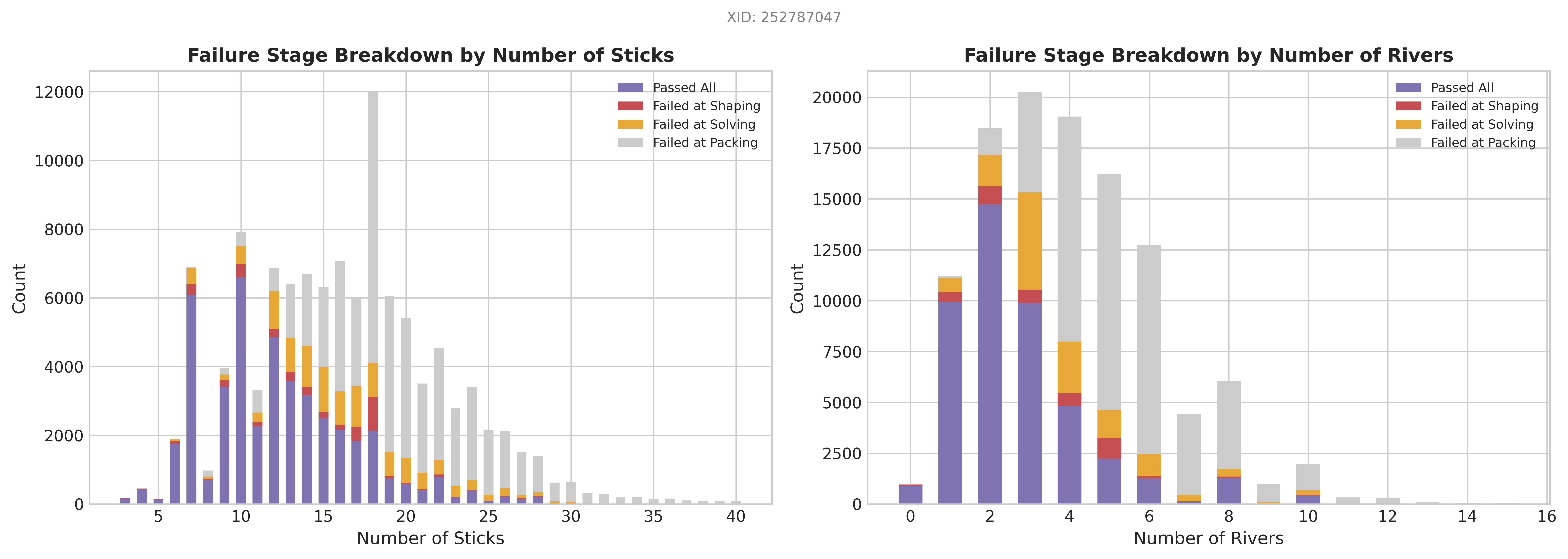}
    \caption{Failure stage breakdown by number of sticks (left) and number of rivers (right). As structural complexity increases, a larger proportion of designs fail at the packing and solving stages.}
    \label{fig:failure_breakdown}
\end{figure}

The progression of candidate designs through the computational generation pipeline, tracking survival from initial concepts to the final curated dataset is shown in \cref{fig:tq_summary}. Out of 560,000 initial tree candidates, 113,276 successfully generate valid semantic stick figures (a 20.2\% pass rate). Subsequent sequential phases filter candidates based on discrete base packing feasibility (55.3\% pass rate), deterministic solving for flat-foldability (79.2\% pass rate), and algorithmic 3D shaping (92.0\% pass rate). Finally, models undergo rigorous verification via simulation strain checks and Vision-Language Model (VLM) aesthetic evaluation; 17,789 designs are filtered out during this stage (7,490 due to low VLM correspondence rewards and 10,299 failing the tree similarity threshold). The completed first-stage pipeline yields 27,869 structurally viable and visually compelling baseline models, representing an overall survival rate of 5.0\% to serve as the foundation for downstream reinforcement learning. To understand the performance across different subjects, we evaluate the success rates broken down by semantic category as shown in \cref{fig:sr_by_category} in the appendix, ordered by the proportion of models that successfully pass all stages of the pipeline. \cref{fig:failure_breakdown} illustrates how the structural complexity of the input stick figure—measured by the number of flaps and rivers—impacts the failure distribution.

\begin{figure*}
    \centering
    \includegraphics[width=\textwidth]{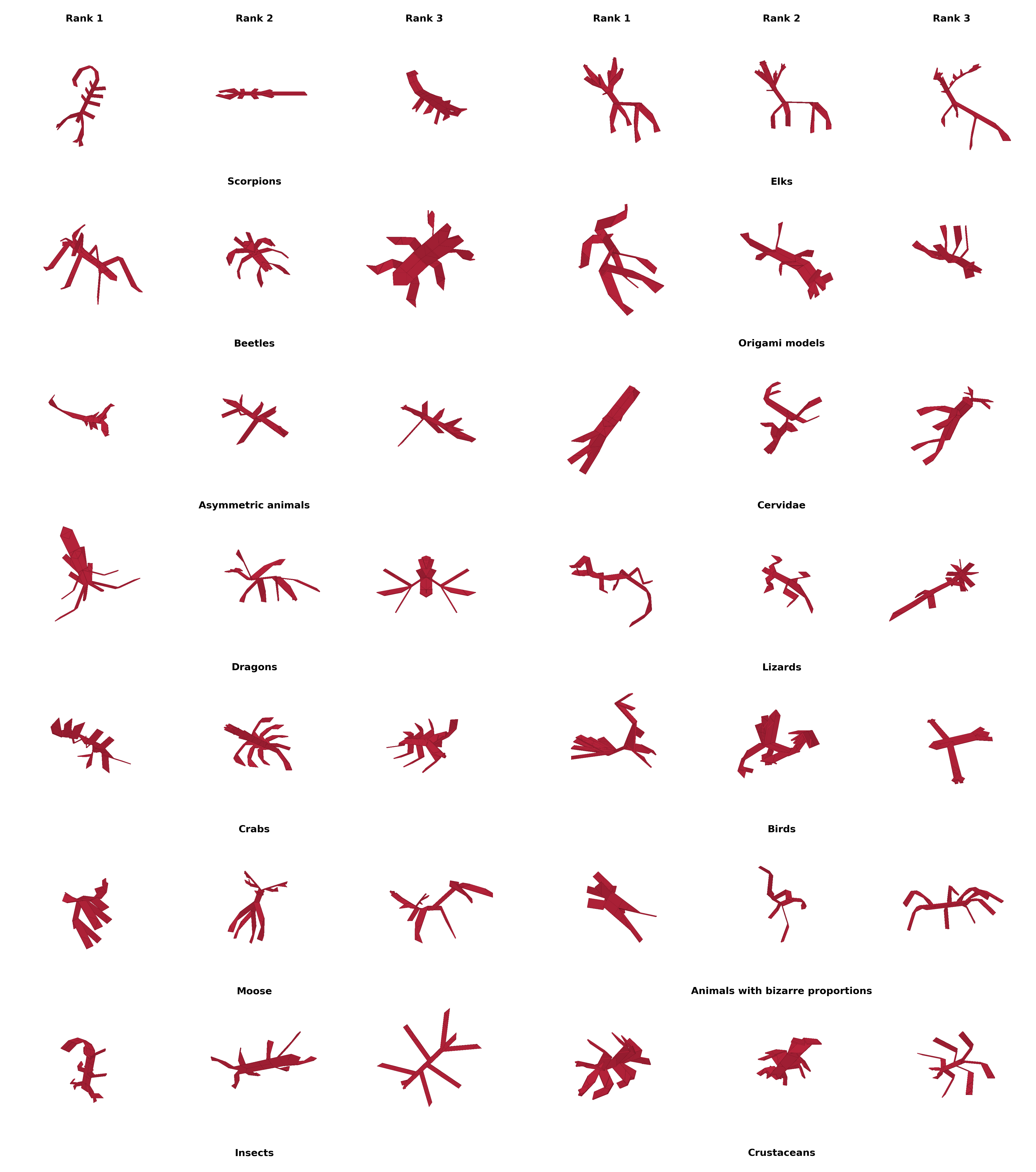}
    \caption{Top three tournament winners per category. The semantic categories (row labels) were used to sample the target origami subjects, driving the initial stick figure generation (\cref{subsec:sf}).}
    \label{fig:tq_winners}
\end{figure*}

To evaluate and select the highest-quality generated origami designs, we implement a distributed, multi-phase Vision-Language Model (VLM) tournament that evaluates folded models using parallelized pairwise VLM comparisons (the Comparison Judge Mode described in \cref{sec:vlm_feedback}). To balance scalability, representation quality, and thematic diversity, the tournament proceeds in four distinct stages: first, a localized Swiss-system tournament is run for each origami model to select the single most representative angle among multiple rendered views; second, independent Swiss-system local tournaments run in parallel within each semantic category to identify local winners; and third, these regional victors compete in a final global Swiss-system tournament to establish an overall quality ranking. Each tournament runs for $\log_2(n)$ rounds (rounded up), where $n$ is the number of tournament candidates (e.g., 3 rounds for 7-view selection). The global results are then subjected to a diversity filter that enforces location-specific caps per super-category, preventing any single origami model from dominant representation among the final top-N winners. \cref{fig:tq_winners} presents the top-3 winners in each category at the end of this stage. 

\newpage
\clearpage
\break

\subsection{Reinforcement Learning}
\begin{figure}[h]
    \centering
    \includegraphics[width=0.9\columnwidth]{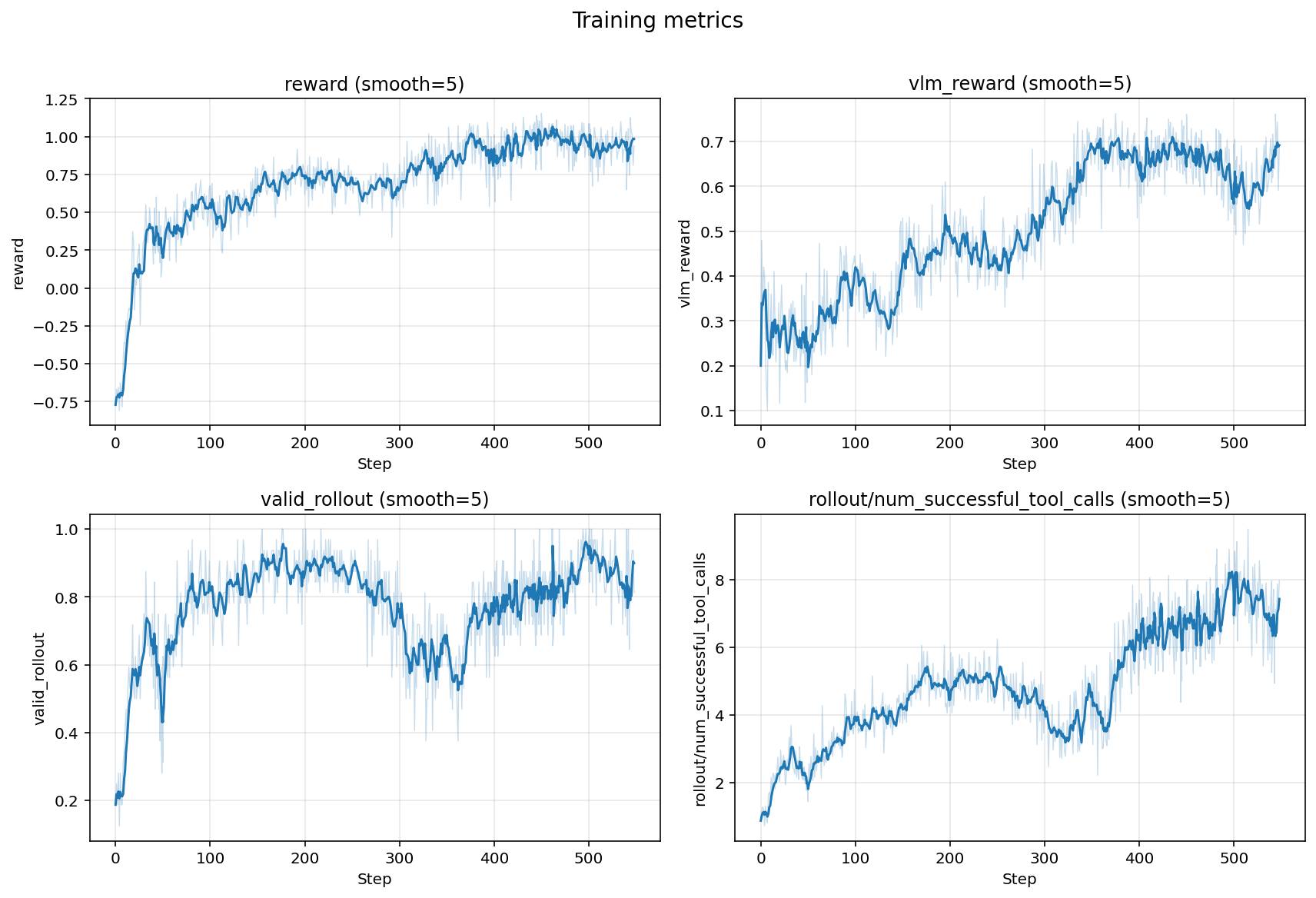}
    \caption{Core RL metrics vs. training steps.}
    \label{fig:rl_metrics}
\end{figure}

The subsequent RL phase is engineered to systematically explore a more expansive and expressive morphological design space, operating strictly from the structurally validated base crease patterns established in the first stage (top-1000 winners). During this phase, the agent's action space is broadened to incorporate advanced geometric shaping tools, such as structural narrowing and more flexible simple folding. This dual-reward structure effectively prevents convergence on local optima, driving the system to discover highly creative, visually compelling, and structurally intricate 3D configurations. \cref{fig:rl_metrics} demonstrates how the key RL metrics increase during training, including the number of successful shaping actions, the VLM reward, the valid rollout percentage and the reward. 

\textbf{Setup.} The RL run involved a batch size of $64$ and a learning rate of $10^{-4}$. The training followed a simple policy gradient algorithm with KL distance towards the base policy. The weight of the KL coefficient was decayed from 1 to $10^{-4}$ over $500$ steps.

\begin{figure}[h]
    \centering
    \includegraphics[width=\columnwidth]{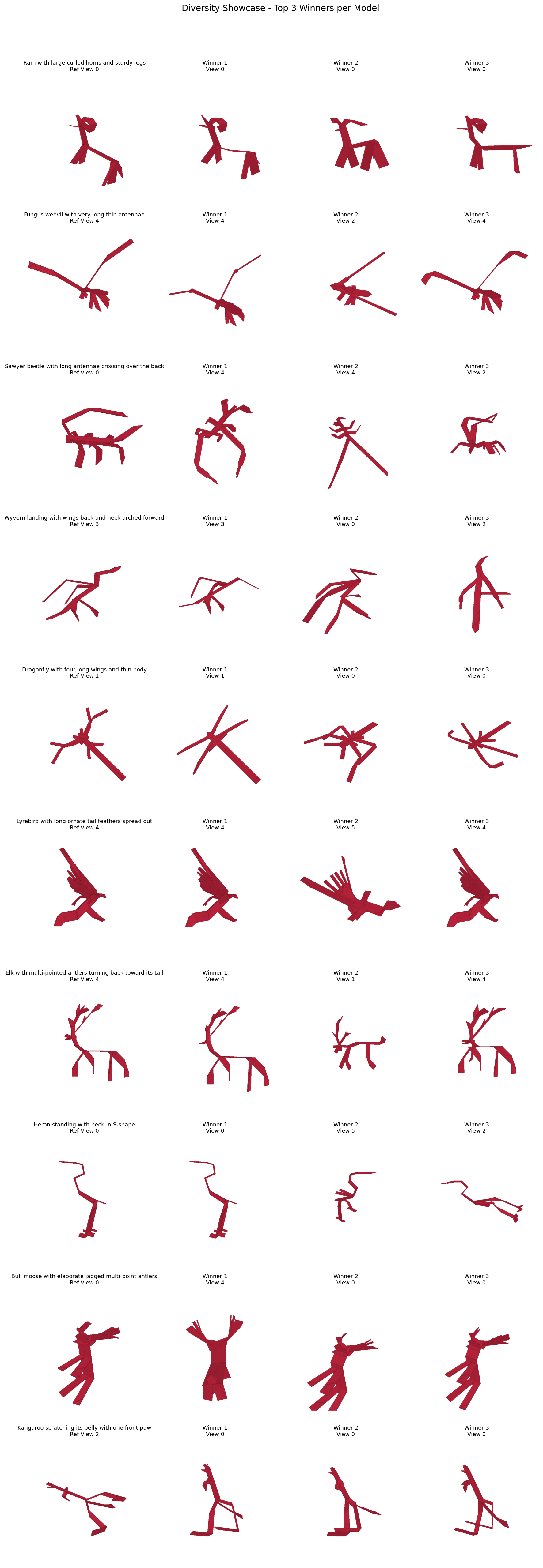}
    \caption{Shaping source origami models in different ways using RL}
    \label{fig:rl_diversity}
\end{figure}

All of the samples that were produced during the RL stage were then sent to another tournament. \cref{fig:rl_diversity} showcases the top-10 winners of that tournament. The source, algorithmic shaping is shown on the image in the left column, while three RL shaping examples (the winners of a local tournament) are shown to the right. As we can see, this second phase successfully produces shaped models that are different from the source algorithmic model, while maintaining the same topology. We can also notice the use of narrowing techniques in some of the models. 

Finally, to produce \cref{fig:top10}, 200 models from RL were manually selected by our team and were put to a final tournament to select the top-10. The manual selection involved visually inspecting RL samples with high VLM reward and high number of tool calls that are visually recognisable. This is the only figure in this paper where human selection played a role.



\newpage
\clearpage
\break
\section{Discussion}
The intersection of artificial intelligence and mathematics to generate aesthetically pleasing, physically plausible objects represents a critical frontier in computational creativity, demanding that systems simultaneously navigate subjective human aesthetics and rigid physical constraints. To overcome the well-documented limitations of standard generative AI in multi-step spatial reasoning, the COrigami framework 
employs an end-to-end neuro-symbolic pipeline that effectively decouples subjective conceptualisation from mathematically rigid execution. Gemini, with additional RL training, handles the semantic generation and shaping at the beginning and end of the process, while the structural core relies purely on custom algorithmic enhancements to known foldability theorems. Analysed through established theoretical frameworks, this architecture robustly operationalises Simon Colton's "Creative Tripod" \citep{Colton_2008}: it demonstrates skill through algorithmic box-pleating that guarantees geometric flat-foldability, imagination via RL exploration, and appreciation through a multi-perspective Vision-Language Model (VLM) feedback loop acting as an autonomous aesthetic critic. Ultimately, COrigami functions as a powerful collaborative partner, supporting human artists by suggesting structural layouts on the grid for complex design topologies, thereby assisting with initial prototypes while enabling the designer to focus on expressive shaping.

Nevertheless, several challenges remain, as our current framework only scratches the surface of computational origami assistance. Thus far, our focus has been strictly on pure origami operating within established box-pleating paradigms; true transformational creativity in this domain will ultimately stem from inventing entirely novel folding methods rather than merely optimizing existing ones. Furthermore, even within discrete box pleating, our current pipeline relies on a limited set of shaping mechanisms (simple folds and narrowing). For expert origami artists, the intermediate, 'rough' 3D shape generated by our pipeline acts as a valuable structural blueprint—a mathematically sound canvas that they can take over to apply their own expressive shaping styles. To expand the computational capabilities of the pipeline, future iterations should incorporate more advanced structural routing mechanisms, such as Pythagorean stretches or level shifters, which human experts frequently utilize to achieve superior packing efficiencies on the grid. While these non-orthogonal elements have historically been restricted to interactive editors because they complicate automated tiling (see Section~\ref{sec:related_work}), integrating them into our backtracking packer would represent a significant step forward in automated layout design. More broadly, while our primary focus was to establish the first fully automated, end-to-end pipeline, we acknowledge that substantial room for improvement remains across its individual components, providing a strong foundation for future research to expand upon.

Finally, the computational complexity of origami design presents both fascinating insights and ongoing hurdles. The mathematical literature extensively proves that determining whether a general crease pattern can fold flat, or assigning valid mountain-valley orientations, is strictly NP-complete—an intractability that persists even under restricted box-pleating grids due to cascading parity constraints and exponential branching factors. Given this, it was particularly interesting to observe that simple, priority-driven greedy algorithms were surprisingly efficient for fairly complex models. By decomposing the spatial geometry into disjoint, hierarchical partitions and using local heuristic guidance, our solver effectively mitigates this complexity.

Ultimately, however, these greedy algorithms do not scale to more complex topologies. As demonstrated in \cref{fig:failure_breakdown}, algorithmic efficiency is eventually bottlenecked by the structural density of the semantic tree, leading to significantly higher failure rates when attempting to resolve highly complex, densely constrained designs. Future work must address this by integrating more advanced machine learning techniques and robust exploration strategies to reliably conquer the most computationally unforgiving regions of the design space.

\clearpage

\bibliography{refs}

@article{agarwal2026origamibench,
author    = {Naaisha Agarwal and Yihan Wu and Yichang Jian and Yifei Peng and Yao-Xiang Ding and Nishad Mansoor and Yikuan Hu and Mohan Li and Wang-Zhou Dai and Emanuele Sansone},
title     = {OrigamiBench: An Interactive Environment to Synthesize Flat-Foldable Origamis},
journal   = {arXiv preprint arXiv:2603.13856},
year      = {2026}
}

@book{demaine2007geometric,
author    = {Erik D. Demaine and Joseph O'Rourke},
title     = {Geometric Folding Algorithms: Linkages, Origami, Polyhedra},
publisher = {Cambridge University Press},
year      = {2007}
}

@article{justin1986mathematics,
author  = {Jacques Justin},
title   = {Mathematics of origami, part 9},
journal = {British Origami},
pages   = {28--30},
month   = {June},
year    = {1986}
}

@book{kasahara1987origami,
author    = {Kunihiko Kasahara and Toshie Takahama},
title     = {Origami for the Connoisseur},
publisher = {Japan Publications},
address   = {Tokyo, Japan},
year      = {1987}
}

@inproceedings{kawasaki1991relation,
author    = {Toshikazu Kawasaki},
title     = {On the relation between mountain-creases and valley-creases of a flat origami},
booktitle = {Proceedings of the First International Conference on Origami in Education and Therapy (COET '91)},
pages     = {229--237},
year      = {1991},
publisher = {British Origami Society}
}

@inproceedings{Colton_2008,
  title={Creativity Versus the Perception of Creativity in Computational Systems},
  author={Colton, Simon},
  booktitle={AAAI Spring Symposium: Creative Intelligent Systems},
  year={2008}
}

@article{feng2025generating,
  title={Generating Creative Chess Puzzles},
  author={Feng, Xidong and Veeriah, Vivek and Chiam, Marcus and Dennis, Michael and Pachauri, Ryan and Tumiel, Thomas and Barbero, Federico and Obando-Ceron, Johan and Shi, Jiaxin and Singh, Satinder and Hou, Shaobo and Toma{\v{s}}ev, Nenad and Zahavy, Tom},
  journal={arXiv preprint arXiv:2510.23881},
  year={2025}
}

@article{veeriah2025evaluating,
  title={Evaluating In Silico Creativity: An Expert Review of AI Chess Compositions},
  author={Veeriah, Vivek and Barbero, Federico and Chiam, Marcus and Feng, Xidong and Dennis, Michael and Pachauri, Ryan and Tumiel, Thomas and Obando-Ceron, Johan and Shi, Jiaxin and Hou, Shaobo and Singh, Satinder and Toma{\v{s}}ev, Nenad and Zahavy, Tom},
  journal={arXiv preprint arXiv:2510.23772},
  year={2025}
}

@article{stepllm2026,
  title={STEP-LLM: Generating CAD STEP Models from Natural Language with Large Language Models},
  author={Shi, Xiangyu and Ding, Junyang and Zhao, Xu and Zhan, Sinong and Patel, Payal and others},
  journal={arXiv preprint arXiv:2601.12641},
  year={2026}
}

@inproceedings{xing2025llm4svg,
  title={Empowering LLMs to Understand and Generate Complex Vector Graphics},
  author={Xing, Ximing and Hu, Juncheng and Liang, Guotao and Zhang, Jing and Xu, Dong and Yu, Qian},
  booktitle={Proceedings of the IEEE/CVF Conference on Computer Vision and Pattern Recognition (CVPR)},
  year={2025}
}

@inproceedings{yang2025omnisvg,
  title={OmniSVG: A Unified Scalable Vector Graphics Generation Model},
  author={Yang, Yiying and Cheng, Wei and Chen, Sijin and Zeng, Xianfang and Yin, Fukun and Zhang, Jiaxu and Wang, Liao and Yu, Gang and Ma, Xingjun and Jiang, Yu-Gang},
  booktitle={Advances in Neural Information Processing Systems (NeurIPS)},
  year={2025}
}

@incollection{colton2012painting,
  title={The Painting Fool: Stories from Building an Automated Painter},
  author={Colton, Simon},
  booktitle={Computers and Creativity},
  pages={3--38},
  year={2012},
  publisher={Springer}
}

@misc{mitani2005oripa,
  author = {Mitani, Jun},
  title = {{ORIPA}: {O}rigami Pattern Editor},
  year = {2005},
  howpublished = {\url{https://mitani.github.io/oripa/}}
}

@inproceedings{bern1996complexity,
  title={The complexity of flat origami},
  author={Bern, Marshall and Hayes, Barry},
  booktitle={SODA},
  volume={96},
  pages={175--183},
  year={1996},
  organization={Atlanta, GA}
}

@inproceedings{ghassaei2018fast,
  title={Fast, interactive origami simulation using {GPU}-accelerated mass-spring systems},
  author={Ghassaei, Amanda and Demaine, Erik D. and Gershenfeld, Neil},
  booktitle={Origami 7: Proceedings of the 7th International Meeting on Origami in Science, Mathematics, and Education (OSME 2018)},
  volume={3},
  pages={1161--1174},
  year={2018}
}

@article{colton2012computational,
  title={Computational Creativity: The Coming of Age},
  author={Colton, Simon and Wiggins, Geraint A},
  journal={AI Magazine},
  volume={33},
  number={3},
  pages={91--91},
  year={2012}
}

@article{pun2025generating,
  title={Generating Physically Stable and Buildable Brick Structures from Text},
  author={Pun, Ava and Deng, Kangle and Liu, Ruixuan and Ramanan, Deva and Liu, Changliu and Zhu, Jun-Yan},
  journal={arXiv preprint arXiv:2505.05469},
  year={2025}
}

@book{lang2012origami,
  title={Origami design secrets: mathematical methods for an ancient art},
  author={Lang, Robert J},
  year={2012},
  publisher={CRC Press}
}

@inproceedings{lang1996treemaker,
author    = {Robert J. Lang},
title     = {A computational algorithm for origami design},
booktitle = {Proceedings of the Twelfth Annual Symposium on Computational Geometry (SoCG '96)},
pages     = {98--105},
year      = {1996},
publisher = {ACM}
}

@article{Akitayaetal2017,
  author  = {Akitaya, Hugo A. and D. Demaine, Erik and S. Ku, Jason},
  title   = {Simple Folding is Really Hard},
  journal = {Journal of Information Processing},
  year    = {2017},
  volume  = {25},
  pages   = {580--589},
  doi     = {10.2197/ipsjjip.25.580}
}

@article{demaine2017origamizer,
  title={Origamizer: A practical algorithm for folding any polyhedron},
  author={Demaine, Erik and Tachi, Tomohiro},
  journal={33rd International Symposium on Computational Geometry (SoCG)},
  year={2017}
}

@article{banarse2026evolution,
  title={Evolution \& Foundation: AI Shares Creative Control},
  author={Banarse, Dylan and Todd, Stephen and Latham, William and Leymarie, Frederic Fol},
  journal={arXiv preprint arXiv:2606.16849},
  year={2026}
}

@article{Arkinetal2000,
  author  = {Arkin, Esther M. and Bender, Michael A. and Demaine, Erik D. and Demaine, Martin L. and Mitchell, Joseph S. B. and Sethia, Saurabh and Skiena, Steven S.},
  title   = {When Can You Fold a Map?},
  journal = {arXiv},
  year    = {2000},
  doi     = {10.48550/arxiv.cs/0011026}
}

@article{Lietal2025,
  author  = {Li, Zhong-Zhi and Zhang, Duzhen and Zhang, Ming-Liang and Zhang, Jiaxin and Liu, Zengyan and Yao, Yuxuan and Xu, Haotian and Zheng, Junhao and Wang, Pei-Jie and Chen, Xiuyi and others},
  title   = {From System 1 to System 2: A Survey of Reasoning Large Language Models},
  journal = {arXiv},
  year    = {2025},
  doi     = {10.48550/arxiv.2502.17419}
}

@article{Bhatietal2026,
  author  = {Bhati, H. and others},
  title   = {Agentic AI in the Software Development Lifecycle},
  journal = {arXiv},
  year    = {2026}
}

@article{guo2025deepseek,
  title={DeepSeek-R1 incentivizes reasoning in LLMs through reinforcement learning},
  author={Guo, Daya and Yang, Dejian and Zhang, Haowei and Song, Junxiao and Wang, Peiyi and Zhu, Qihao and Xu, Runxin and Zhang, Ruoyu and Ma, Shirong and Bi, Xiao and others},
  journal={Nature},
  volume={645},
  number={8081},
  pages={633--638},
  year={2025},
  publisher={Nature Publishing Group UK London}
}

@article{Demaineetal2015,
  author  = {Demaine, Erik D. and Eppstein, David and Hesterberg, Adam and Ito, Hiro and Lubiw, Anna and Uehara, Ryuhei and Uno, Yushi},
  title   = {Folding a Paper Strip to Minimize Thickness},
  journal = {Lecture Notes in Computer Science},
  year    = {2015},
  pages   = {113--124},
  doi     = {10.1007/978-3-319-15612-5_11}
}

@article{Hull2023,
  author  = {Hull, Thomas C. and Zakharevich, Inna},
  title   = {Flat origami is Turing Complete},
  journal = {arXiv},
  year    = {2023},
  doi     = {10.48550/arxiv.2309.07932}
}

@misc{tsai2020boxpleating,
author       = {Mu-Tsun Tsai},
title        = {Box Pleating Studio: Super-complex origami design made easy!},
howpublished = {\url{https://bpstudio.abstreamace.com/}},
year         = {2020}
}

@inproceedings{xu2025origamispace,
author    = {Rui Xu and Dakuan Lu and Zicheng Zhao and Xiaoyu Tan and Xintao Wang and Siyu Yuan and Jiangjie Chen and Yinghui Xu},
title     = {OrigamiSpace: Benchmarking Multimodal LLMs in Multi-Step Spatial Reasoning with Mathematical Constraints},
booktitle = {Advances in Neural Information Processing Systems (NeurIPS)},
year      = {2025}
}

@inproceedings{FlatFolder_OSME2024,
  author    = {Hugo A. Akitaya and Erik D. Demaine and Jason S. Ku},
  title     = {Computing Flat-Folded States},
  booktitle = {Origami$^8$: Proceedings of the 8th International Meeting on Origami in Science, Mathematics and Education (OSME 2024)},
  address   = {Melbourne, Australia},
  month     = {July 16--18},
  year      = {2024},
  pages     = {to appear}
}

@inproceedings{Demaine1999,
  author    = {Demaine, Erik D. and Demaine, Martin L. and Lubiw, Anna},
  title     = {Folding and one straight cut suffice},
  booktitle = {Proceedings of the 10th Annual ACM-SIAM Symposium on Discrete Algorithms (SODA '99)},
  year      = {1999},
  pages     = {891--892}
}

@inproceedings{Demaine1998,
  author    = {Demaine, Erik D. and Demaine, Martin L. and Lubiw, Anna},
  title     = {Folding and cutting paper},
  booktitle = {Revised Papers from the Japan Conference on Discrete and Computational Geometry (JCDCG '98)},
  year      = {1998},
  pages     = {104--117},
  publisher = {Springer}
}

@inproceedings{oh2015dissimilarity,
  title={A Dissimilarity Measure for Comparing Origami Crease Patterns.},
  author={Oh, Seung Man and Toussaint, Godfried T and Demaine, Erik D and Demaine, Martin L},
  booktitle={ICPRAM (1)},
  pages={386--393},
  year={2015}
}

@techreport{GeminiTeam2025Gemini3,
  author      = {{Gemini Team}},
  title       = {Gemini 3 Model Card},
  institution = {Google DeepMind},
  year        = {2025}
}

@article{comanici2025gemini,
  title={Gemini 2.5: Pushing the frontier with advanced reasoning, multimodality, long context, and next generation agentic capabilities},
  author={Comanici, Gheorghe and Bieber, Eric and Schaekermann, Mike and Pasupat, Ice and Sachdeva, Noveen and Dhillon, Inderjit and Blistein, Marcel and Ram, Ori and Zhang, Dan and Rosen, Evan and others},
  journal={arXiv preprint arXiv:2507.06261},
  year={2025}
}

@article{LangAlperin2014,
  title={Graph paper for polygon-packed origami design},
  author={Lang, Robert J and Alperin, Roger C},
  journal={Origami6: I. Mathematics},
  pages={305--318},
  year={2014}
}

@article{hubert2025olympiad,
title	= {Olympiad-Level Formal Mathematical Reasoning with Reinforcement Learning},
author	= {Thomas Hubert and Rishi Mehta and Laurent Sartran and Miklós Z. Horváth and Goran Žužić and Eric Wieser and Aja Huang and Julian Schrittwieser and Yannick Schroecker and Hussain Masoom and Ottavia Bertolli and Tom Zahavy and Amol Mandhane and Jessica Yung and Iuliya Beloshapka and Borja Ibarz and Vivek Veeriah and Lei Yu and Oliver Nash and Paul Lezeau and Salvatore Mercuri and Calle Sönne and Bhavik Mehta and Alex Davies and Daniel Zheng and Fabian Pedregosa and Yin Li and Ingrid von Glehn and Mark Rowland and Samuel Albanie and Ameya Velingker and Simon Schmitt and Edward Lockhart and Henryk Michalewski and Nicolas Sonnerat and Demis Hassabis and Pushmeet Kohli and David Silver},
year	= {2025},
URL	= {https://www.nature.com/articles/s41586-025-09833-y},
journal	= {Nature}}

@inproceedings{justin1997towards,
  title={Towards a mathematical theory of origami},
  author={Justin, Jacques},
  booktitle={Proceedings of the Second International Meeting of Origami Science and Scientific Origami, 1997},
  year={1997}
}

@ARTICLE{schmidhuber2010creativity,
  author={Schmidhuber, Jürgen},
  journal={IEEE Transactions on Autonomous Mental Development}, 
  title={Formal Theory of Creativity, Fun, and Intrinsic Motivation (1990–2010)}, 
  year={2010},
  volume={2},
  number={3},
  pages={230-247},
  doi={10.1109/TAMD.2010.2056368}}

\newpage
\appendix
\onecolumn

\section{Related Work}
\label{sec:related_work}
The computational treatment of origami has undergone a profound evolution over the past four decades, transitioning from the formalization of local geometric axioms to the development of sophisticated, neuro-symbolic artificial intelligence systems capable of end-to-end design. This section reviews the foundational theorems of foldability, the shift from continuous to discrete generative optimization, and the current state-of-the-art in multimodal spatial reasoning.

\subsection*{Axiomatic Foundations and Local Flat-Foldability}
The analytical verification of flat-foldability begins with the geometric conditions of an isolated vertex. Kawasaki's Theorem establishes that a vertex can fold flat if and only if the alternating sum of its incident sector angles equals precisely 180 degrees \citep{kawasaki1991relation}. While this addresses the continuous angular constraints, the combinatorial assignment of fold orientations—Mountain ($M$) or Valley ($V$)—is governed by Maekawa's Theorem, which mandates that $M - V = \pm 2$ at any flat-foldable interior vertex \citep{kasahara1987origami}. These parities were simultaneously and independently explored across the globe in the late 1980s, notably by Jacques Justin, whose mathematical characterizations challenged contemporary assumptions about ruled surfaces in paper folding \citep{justin1986mathematics}. In modern algorithmic implementations, these static constraints are often supplemented by procedural ``crimping'' simulations to evaluate local sector bounds and prevent layer penetrations dynamically.

\subsection*{Global Flat-Foldability and Computational Complexity}
While local flat-foldability is trivially verified, guaranteeing global flat-foldability—ensuring a crease pattern physically folds without continuous self-intersection—is strictly NP-hard. Historically, verification relied on a ``pointwise'' definition, necessitating the intractable geometric checking of infinite sets of points across a continuous manifold \citep{demaine2007geometric}. A fundamental paradigm shift occurred with Akitaya et al.'s formulation of the ``facewise'' definition \citep{FlatFolder_OSME2024}. By transforming the problem into a finite constraint-satisfaction graph tracking discrete overlapping faces (e.g., Taco-Taco, Taco-Tortilla constraints), this approach allows valid layer-ordering to be computed in polynomial $O(n^3)$ time, drastically accelerating physical simulations.

\subsection*{Generative Design: From Continuous to Discrete Frameworks}
The inverse problem—generating a crease pattern from a target shape—was formalized by Lang's ``Tree Method,'' which mathematically framed the process as an interwoven circle-packing optimization \citep{lang1996treemaker}. Similarly, Origamizer \citep{demaine2017origamizer} approached the generative problem by mapping arbitrary 3D polyhedral meshes to flat-foldable crease patterns. While this continuous spatial optimization optimally minimized square paper dimensions, it generated arbitrary irrational reference points that presented severe physical execution challenges for folders.

Consequently, modern computational design relies heavily on ``box pleating'' (BP), which restricts all structural creases to an orthogonal integer grid intersected by 45-degree diagonal ridges. Though algorithmic box pleating is NP-hard due to cascading parity constraints, systems like Tsai's Box Pleating Studio \citep{tsai2020boxpleating} introduced advanced routing mechanisms (e.g., Generalized Offset Pythagorean Stretches) to achieve high packing efficiencies on discrete grids. This discretization is critical, as it transforms continuous spatial puzzles into tractable combinatorial state-space searches.

In developing COrigami, we explored integrating both tools into our pipeline. We automated TreeMaker's core optimization, loop—scale optimization, crease pattern construction, and edge strain minimization—by writing a programmatic interface that ingests our stick figures and runs these steps without manual intervention. However, many complex topologies failed to produce valid crease patterns because several critical interactive steps were not automated: imposing symmetry conditions by pairing nodes about a symmetry line, fracturing high-order polygons by adding auxiliary stubs, forcing leaf nodes to paper edges to reduce layer count, and manually repositioning nodes to escape local optima. We also investigated BP Studio's packing solver, which operates on the discrete grid but solves a continuous relaxation of the packing problem. As a result, its solutions frequently contain Pythagorean stretches and cannot enforce contiguous tiling—the elimination of all empty space on the sheet—which is a prerequisite for generating a valid base crease pattern. A hybrid workflow combining TreeMaker's initial packing with BP Studio's discrete optimization encountered the same limitations.

These practical barriers motivated our decision to build an integrated backtracking packer on a strictly orthogonal box-pleating grid, yielding designs that are easier to understand, fold, and verify by hand. Unlike traditional computational tools such as TreeMaker and BP Studio, which explicitly optimize for packing efficiency, COrigami optimizes for success rate and visual recognizability at the cost of efficiency. Taking this approach to the extreme, naive 'stacking' algorithms—where flaps are packed vertically in traversal order with full grid width, and rivers are placed via contour-hugging around the already-placed flaps (also referred to as Kawahata's ``string-of-beads'' algorithm by Robert Lang in the TreeMaker manual)—empirically packed 100\% of our stick figures. However, this approach is highly inefficient in its use of paper area, often resulting in grid sizes well over 100 even for simple models, which leads to prohibitively dense physical folds.

COrigami navigates between these extremes by generating a diverse manifold of valid, discrete packing solutions across various grid resolutions. By starting from the heuristic lower bound, the system guarantees a baseline level of efficiency. However, rather than selecting the final layout based on mathematical scale or area utilization, the system defers to the Vision-Language Model (VLM) in the single model evaluation mode. This allows the VLM to autonomously filter out visually poor or overly bulky layouts, selecting the packing solution that ultimately yields the highest aesthetic and structural quality. Future work may explore other trade-offs between these objectives. 

\begin{table}[h]
  \centering
  \begin{tabular}{l r r}
    \toprule
    \textbf{Outcome} & \textbf{Count} & \textbf{\%} \\
    \midrule
    Success & 743 & 1.1 \\
    \midrule
    \multicolumn{3}{l}{\textit{Failure breakdown}} \\
    \quad CP construction failed & 49,120 & 74.4 \\
    \quad Scale optimisation failed & 7,012 & 10.6 \\
    \quad Flat-foldability violation & 8,736 & 13.2 \\
    \quad Timeout & 425 & 0.6 \\
    \midrule
    \textbf{Total} & \textbf{66,036} & \\
    \bottomrule
  \end{tabular}
  \caption{TreeMaker results on 66,036 stick figures. ``CP construction failed'' groups cases where crease pattern building failed both before and after edge strain optimisation. ``Flat-foldability violation'' includes patterns that were built but failed our flat-foldability verification.}
  \label{tab:treemaker}
\end{table}

It should be noted that because TreeMaker relies heavily on human-in-the-loop interactions to resolve complex topologies, its performance in this fully automated benchmark likely under represents its efficacy, as critical manual interventions were bypassed to enable programmatic evaluation.

Lastly, beyond the standard orthogonal grids used in box pleating, alternative grid systems that provide similar geometric guarantees have been theoretically explored. For instance, "hex pleating" restricts creases to a hexagonal grid (multiples of 30°), which similarly ensures both rational folding angles and the finite propagation of creases. Although some origami artists have manually designed modules using hex pleating, no computational model has yet been developed for it due to a lack of practical demand. More broadly, \citet{LangAlperin2014} demonstrated that there exists a countable infinity of such grids capable of providing angle-constrained creases and forcing finite crease propagation. However, the vast majority of these grids offer no distinct geometric advantages and remain highly impractical for actual origami design, reinforcing the orthogonal grid as the standard for discrete computational frameworks like COrigami.

\subsection*{Spatial Intelligence and Neuro-Symbolic Pipelines}
The emergence of Multimodal Large Language Models (MLLMs) introduced potential for computational design assistance; however, deep neural networks fundamentally struggle with invariant geometric properties. The \textit{OrigamiSpace} benchmark \citep{xu2025origamispace} and the \textit{OrigamiBench} environment \citep{agarwal2026origamibench} empirically exposed severe deficits in modern models (like GPT-4o and Gemini 2.5) concerning multi-step spatial reasoning and structural violations (e.g., paper self-intersection).

\subsection*{Origami Software and Editors}
Alongside the algorithmic generators, dedicated drawing and editing tools were developed to digitize and test discrete crease patterns. A foundational tool in this space is \href{https://mitani.cs.tsukuba.ac.jp/origami_application/}{ORIPA (Origami Pattern Editor)}, created by Jun Mitani at the University of Tsukuba in 2005. ORIPA, a Java-based application, allows users to draw crease patterns while maintaining strict line typologies (1 for Contour, 2 for Mountain, 3 for Valley) and exports these designs into custom \texttt{.opx} or standard ASCII \texttt{.cp} files. Crucially, ORIPA features an embedded estimation engine capable of calculating and rendering the folded shape directly from the flat-foldable crease pattern. 

Building upon this foundation of static pattern digitisation, more recent tools have shifted focus toward dynamic physical evaluation. The \href{https://origamisimulator.org/}{Origami Simulator}, developed by Amanda Ghassaei, Erik Demaine, and Neil Gershenfeld introduced a highly parallelised, GPU-accelerated WebGL application for fast, interactive origami simulation. Rather than calculating sequential, rigid folding steps, the simulator evaluates the entire crease pattern simultaneously by iteratively solving for small geometric displacements caused by crease forces across the sheet. This real-time feedback loop—coupled with strain and deformation visualisations—provides an accessible method for designers to evaluate the physical viability of their complex layouts before execution.

However, a fundamental limitation of all current computational folding engines—including ORIPA and Origami Simulator—is their reliance on a \emph{zero-thickness paper assumption} that treats the sheet as an idealized 2D manifold. In physical folding, however, paper possesses a finite, non-zero thickness ($t > 0$). In dense box-pleated layouts, paper layers stack repeatedly; a single appendage can accumulate dozens of overlapping sheets, creating severe layer accumulation (bulking) that requires substantial folding force and compresses the fibers~\citep{Demaineetal2015}. This physical bulk introduces "paper creep" (thickness shift), forcing outer layers of paper to wrap around inner layers and geometrically displacing crease lines from their idealized grid coordinates. Because zero-thickness simulators cannot model these physical bulking stresses, paper tension, or material-dependent tearing limits, the 3D models and crease patterns generated by COrigami function strictly as \emph{mathematically sound structural starting points}. Realizing these designs requires physical interpretation by a human artist, who must utilize specialized thin mediums (such as double tissue paper or Washi) and tactile shaping expertise (e.g., wet-folding, closed-sink folds, and manual layer-thinning) to resolve physical bulking constraints and successfully execute the final physical form.

\textbf{Creative Generation in Structured Domains}
Beyond purely pixel-based image generation, recent work has explored utilizing Large Language Models (LLMs) and Vision-Language Models (VLMs) for creative generation within highly structured and geometrically constrained domains. In computer-aided design (CAD), frameworks such as STEP-LLM \citep{stepllm2026} generate parametric 3D CAD STEP models directly from natural language prompts, leveraging reserialization strategies to preserve graph-like structural logic. In the realm of 2D design, vector graphics synthesis has similarly benefited from tokenizing geometry; models like LLM4SVG \citep{xing2025llm4svg} and OmniSVG \citep{yang2025omnisvg} parameterize Scalable Vector Graphics (SVG) commands and coordinates into discrete tokens, allowing autoregressive models to synthesize complex, editable vector designs while reducing structural occlusion and coordinate hallucinations. Extending these principles to physical 3D assembly, LegoGPT \citep{pun2025generating} represents interconnecting brick structures as tokenized text sequences to generate LEGO designs from natural language prompts, employing a physics-aware rollback and structural validity check during autoregressive inference to guarantee physical stability and buildability. This challenge of generating valid yet highly creative content within mathematically rigid environments is not unique to spatial design, extending also to abstract board games. In chess, \citet{feng2025generating} introduced a reinforcement learning framework that leverages chess engine search statistics as reward signals to generate novel, counter-intuitive chess puzzles, while \citet{veeriah2025evaluating} conducted an extensive expert evaluation of these "in silico" compositions to study human-AI alignment in aesthetic domain judgment. 

Beyond rigid geometric constraints, generative and evolutionary AI systems have also been utilized to steer creative processes in organic digital art \citep{banarse2026evolution}, which was showcased at the \href{https://www.evolutionandfoundation.com/}{"Evolution and Foundational AI" exhibition}. This approach integrates a legacy 3D form-growing grammar with the multimodal visual reasoning capabilities of a frontier foundation model (Google Gemini) acting as an automated aesthetic curator. Rather than relying on human intervention for every selective breeding step across an expansive genetic space, their system leverages the VLM to perform binary tournaments and interpret semantic archetypes within emergent, abstract 3D phenotypes. This architectural setup shifts the human creator's role from an intensive manual curator to an overarching system designer. In a similar vein, our framework offloads low-level geometric and topological curation to specialized algorithmic and tokenized learning blocks, allowing the artist to focus entirely on high-level collaborative direction.


These diverse, multi-objective pipelines align closely with theoretical frameworks established within the field of computational creativity. Specifically, Simon Colton's "Creative Tripod" formulation \citep{Colton_2008} posits that a system must exhibit skill, imagination, and appreciation to be perceived as genuinely creative—criteria historically investigated in autonomous art-generation systems like The Painting Fool \citep{colton2012painting}. By combining systematic geometric generation (skill) with vast structural exploration (imagination) and automated aesthetic criticism (appreciation), modern frameworks operationalize these core dimensions, helping computational creativity mature as an independent and rigorous scientific discipline \citep{colton2012computational}. Together, these developments point to a growing paradigm of combining large generative models with rigid symbolic structures to achieve verifiable, physically reproducible, and aesthetically compelling creative outputs.

\section{Generating crease patterns in SVG space}
\label{sec:cp_sft_rl}
\begin{figure}[h]
    \centering
    \includegraphics[width=\linewidth]{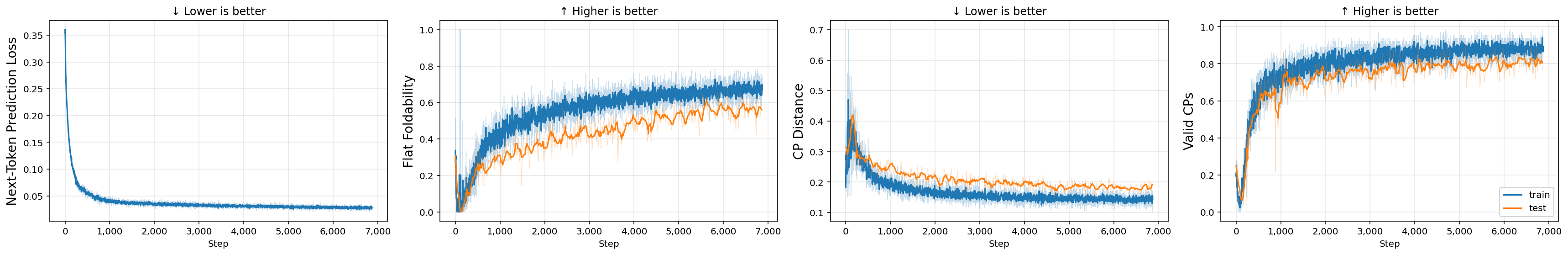}
    \caption{Fine-tuning Gemini model on synthetic crease-pattern datasets.}
    \label{fig:sft}
\end{figure}

To establish a baseline for unconstrained generative architectures, we investigated whether a language model could be directly fine-tuned to produce valid crease patterns. We trained a Gemini model on 400k synthetic crease patterns (approximately 3.2B tokens). This data was generated via a highly scalable TreeMaker-based pipeline that produced diverse, flat-foldable patterns, though it could not generate visually recognizable origami designs. As shown in \cref{fig:sft}, fine-tuning Gemini yields rapid initial learning progress. Structural syntax validity and mathematical flat-foldability clearly improve during early training steps, accompanied by a corresponding decrease in policy text loss and crease pattern distance \citep{oh2015dissimilarity}. The model also demonstrates generalisation on a held-out dataset.

Nevertheless, these performance metrics ultimately saturate well below perfection; flat foldability plateaus near 60\% on the test set, demonstrating that the model never reliably achieves fully flat-foldable crease patterns. This hard ceiling is a direct consequence of the compounding generative challenges inherent to the origami domain. Generating a crease pattern for a visually recognisable model requires outputting a highly complex graph containing thousands of shaping creases. Because each individual crease requires tens of tokens to define, the resulting output sequence is exceptionally long. Within this dense topological framework, even minuscule numerical hallucinations or a single-token error cascade into severe flat-foldability violations. These saturation limits empirically validate conclusions recently drawn by spatial intelligence benchmarks, such as OrigamiSpace \citep{xu2025origamispace} and OrigamiBench \citep{agarwal2026origamibench}, which exposed severe deficits in unconstrained multimodal models when navigating invariant geometric properties and multi-step spatial reasoning. This algorithmic ceiling is further exacerbated by the severe scarcity of high-quality training data, as real-world crease patterns traditionally serve only as abstract structural guidelines rather than exhaustive, mathematically rigorous 3D blueprints.

These negative empirical results conclusively demonstrate that unconstrained, end-to-end generation is difficult due to the requirement for strict physical viability, long generation length and lack of training data. This fundamental architectural limitation directly motivated our transition to discrete box pleating. By restricting all axis-parallel creases and hinges strictly to an orthogonal integer grid, we discretise the design space, mapping continuous geometric packing to tractable combinatorial state-space searches. This neuro-symbolic shift allows us to offload the mathematically rigid constraints of global flat foldability to custom deterministic solvers, guaranteeing physical reproducibility while freeing the AI to focus purely on semantic conceptualisation and heuristic shaping.

\section{Stick figure generation}

In this section, we provide further implementation details.

\subsection{Category and Example Generation}
When generating synthetic data, we use a hierarchical prompting approach. First, we prompt the model to generate broad object categories.

\vspace{1em}
\begin{tcolorbox}[
    colback=gray!5!white, 
    colframe=gray!75!black, 
    arc=4pt, 
    boxrule=0.5pt,
    title=Category Generation Prompt,
    fonttitle=\bfseries,
    coltitle=white
]
Generate a diverse list of categories for objects that are simple, and easily represented by an acyclic, stick-figure-like "skeleton". \\[0.5em]
The objects should be represented by simple origami designs. \\[0.5em]
Bad examples include things that are simple solid shapes (like cups or bricks) or have no defined structure (like clouds or water). \\[0.5em]
Think of at least \{ \texttt{number} \} new categories as a JSON list with just 1 or 2 words for each category.
\end{tcolorbox}
\vspace{1em}

Next, we prompt the model to suggest many objects for each category. We require objects to be acyclic and representable by a tree-like skeleton and simple origami. This prevents the generation of unstructured entities (e.g., water, clouds) or solid shapes (e.g., bricks, cups).

\vspace{1em}
\begin{tcolorbox}[
    colback=gray!5!white, 
    colframe=gray!75!black, 
    arc=4pt, 
    boxrule=0.5pt,
    title=Object Generation Prompt,
    fonttitle=\bfseries,
    coltitle=white
]
Make a diverse list of physical objects that have an acyclic, stick-figure-like "skeleton", within the category '\{ \texttt{category} \}'. \\[0.5em]
A good object is one that can be represented by a simple origami design. \\[0.5em]
The objects should be understood by a child. Please avoid imaginary objects, such as yoga poses that do not exist in real life. \\[0.5em]
Good examples have a distinct structure like: \{ \texttt{examples} \}. \\[0.5em]
Bad examples are things that are simple solid shapes (like cups or bricks) or have no defined structure (like clouds or water) or where we cannot reach the required level of detail. \\[0.5em]
Create a JSON list of at least \{ \texttt{number} \} new objects with up to 10 words for each object.
\end{tcolorbox}
\vspace{1em}

Further, we provide Gemini with an example of a stick figure such as:

\vspace{1em}
\begin{tcolorbox}[
    colback=gray!5!white, 
    colframe=gray!75!black, 
    arc=4pt, 
    boxrule=0.5pt,
    title=Stick figure Example,
    fonttitle=\bfseries,
    coltitle=white,
    breakable 
]
{\ttfamily
\{ \\
\hspace*{1em}"name": "cat", \\
\hspace*{1em}"complexity": 1, \\
\hspace*{1em}"color\_top": [255, 196, 0], \\
\hspace*{1em}"color\_bottom": [255, 196, 0], \\
\hspace*{1em}"category": "domestic animal", \\
\hspace*{1em}"root": "body", \\
\hspace*{1em}"children": [ \\
\hspace*{2em}\{ \\
\hspace*{3em}"name": "body", \\
\hspace*{3em}"start": "hips", \\
\hspace*{3em}"end": "shoulders", \\
\hspace*{3em}"length": 4, \\
\hspace*{3em}"azimuth\_angle": 0.0, \\
\hspace*{3em}"elevation\_angle": 0.0 \\
\hspace*{2em}\}, \\
\hspace*{2em}\{ \\
\hspace*{3em}"name": "head", \\
\hspace*{3em}"start": "shoulders", \\
\hspace*{3em}"end": "nose", \\
\hspace*{3em}"length": 1, \\
\hspace*{3em}"azimuth\_angle": 0.0, \\
\hspace*{3em}"elevation\_angle": 45.0 \\
\hspace*{2em}\}, \\
\hspace*{2em}\{ \\
\hspace*{3em}"name": "front\_left\_leg", \\
\hspace*{3em}"start": "shoulders", \\
\hspace*{3em}"end": "front\_left\_paw", \\
\hspace*{3em}"length": 3, \\
\hspace*{3em}"azimuth\_angle": 90.0, \\
\hspace*{3em}"elevation\_angle": -90.0 \\
\hspace*{2em}\}, \\
\hspace*{2em}\{ \\
\hspace*{3em}"name": "front\_right\_leg", \\
\hspace*{3em}"start": "shoulders", \\
\hspace*{3em}"end": "front\_right\_paw", \\
\hspace*{3em}"length": 3, \\
\hspace*{3em}"azimuth\_angle": -90.0, \\
\hspace*{3em}"elevation\_angle": -90.0 \\
\hspace*{2em}\}, \\
\hspace*{2em}\{ \\
\hspace*{3em}"name": "rear\_left\_leg", \\
\hspace*{3em}"start": "hips", \\
\hspace*{3em}"end": "rear\_left\_paw", \\
\hspace*{3em}"length": 3, \\
\hspace*{3em}"azimuth\_angle": 90.0, \\
\hspace*{3em}"elevation\_angle": -90.0 \\
\hspace*{2em}\}, \\
\hspace*{2em}\{ \\
\hspace*{3em}"name": "rear\_right\_leg", \\
\hspace*{3em}"start": "hips", \\
\hspace*{3em}"end": "rear\_right\_paw", \\
\hspace*{3em}"length": 3, \\
\hspace*{3em}"azimuth\_angle": -90.0, \\
\hspace*{3em}"elevation\_angle": -90.0 \\
\hspace*{2em}\}, \\
\hspace*{2em}\{ \\
\hspace*{3em}"name": "tail", \\
\hspace*{3em}"start": "hips", \\
\hspace*{3em}"end": "tail\_tip", \\
\hspace*{3em}"length": 3, \\
\hspace*{3em}"azimuth\_angle": 180.0, \\
\hspace*{3em}"elevation\_angle": 30.0 \\
\hspace*{2em}\} \\
\hspace*{1em}] \\
\}
}
\end{tcolorbox}
\vspace{1em}

\subsection{Tree similarity score}
\label{subsec:tree_sim}
To evaluate how closely a simulated 3D folded origami model matches its target stick-figure configuration, we define a scale-, rotation-, and translation-invariant shape similarity score. To facilitate mapping to the packing solver, we simplify the stick figure by merging linear chains, collapsing straight segments of degree-two joints into single sticks. This merged representation serves as a structural approximation that focuses the evaluation of the 3D shape on leaf extremities and branching junctions while omitting intermediate points. This approximation is sufficient since we only use this score as a coarse filter and tie-breaker and can be improved in future work. Using the 2D crease pattern coordinates derived from the base packing phase, the joints of this merged stick figure are projected into 3D space through the deterministic folding simulator. The 3D coordinates of the common joints in both the target stick figure and the folded model are subsequently centred, normalised to a unit Frobenius norm to achieve scale invariance, and aligned using Procrustes analysis---which leverages Singular Value Decomposition (SVD) to determine the optimal rotation. Finally, the shape similarity is computed from the Procrustes distance $d^2$ as $1 - d^2$ (mathematically equivalent to the squared sum of the singular values $\text{trace}(\Sigma)^2$), yielding a bounded score in the interval $[0, 1]$, where a score of $1.0$ represents a perfect geometric match under rigid transformation.

\section{Flat Foldability}
\label{sec:flat_fold}
Two classical necessary conditions govern local foldability: Kawasaki's theorem requires that the alternating sums of consecutive sector angles at a vertex each equal $180^\circ$~\citep{kawasaki1991relation}, and Maekawa's theorem requires $|M - V| = 2$~\citep{kasahara1987origami, justin1986mathematics}. These conditions are necessary but not sufficient when mountain--valley assignments are considered. For a sufficient local test, we employ a \emph{crimping} algorithm---a generalisation of the Big-Little-Big lemma~\citep{kawasaki1991relation, justin1997towards} that handles equal angles~\citep{demaine2007geometric}. The algorithm maintains a circular list of sector angles ($\theta_i$) and iteratively identifies a non-strict local minimum $\theta_i \leq \theta_{i\pm1}$ whose bounding creases have opposite mountain--valley assignments. It then simulates a crimp: $\theta_i$ and its clockwise neighbour are removed, and the counter-clockwise neighbour is updated to $\theta_{i-1} \leftarrow \theta_{i-1} - \theta_i + \theta_{i+1}$, merging the bounding crease assignments accordingly. This repeats until only two sectors remain; the vertex is flat-foldable if and only if all remaining bounding creases share the same assignment (see~\cref{alg:crimp_angles}).

\begin{algorithm}
\caption{Flat Foldability Check: Crimping}
\label{alg:crimp_angles}
\begin{algorithmic}[1]
\Require A linked list of active sectors $S$ (active = unchecked), where each sector $s \in S$ has an angle $s.\theta$ and bounding creases $s.left, s.right \in \{M, V\}$.
\Ensure \textbf{true} if the vertex is flat-foldable.

\If{$|S| < 2$}
    \State \Return \textbf{false}
\EndIf

\Function{IsLocalMin}{$s$}
    \If{$s.left = s.right$} \Comment{Must have opposite assignments (one M, one V)}
        \State \Return \textbf{false}
    \EndIf
    \State \Return $(s.\theta \leq s.prev.\theta) \land (s.\theta \leq s.next.\theta)$
\EndFunction

\State $Q \gets \{s \in S \mid \Call{IsLocalMin}{s}\}$ \Comment{Initialize set of candidate sectors}

\While{$Q \neq \emptyset$ \textbf{and} $|S| > 2$}
    \State $s \gets Q.\text{pop}()$
    
    \If{$s \notin S$ \textbf{or} \textbf{not} \Call{IsLocalMin}{$s$}}
        \State \textbf{continue}
    \EndIf
    
    \State $L \gets s.prev$
    \State $R \gets s.next$
    
    \State $L.\theta \gets L.\theta - s.\theta + R.\theta$ \Comment{Simulate crimping}
    \State $L.right \gets R.right$ \Comment{Merge bounding creases}
    
    \State $S.\text{remove}(s, R)$ \Comment{Removes $s$ and $R$, updating adjacent pointers}
    
    \For{$x \in \{L.prev, L, L.next\}$} \Comment{Check updated neighborhood}
        \If{\Call{IsLocalMin}{$x$}}
            \State $Q.\text{add}(x)$
        \EndIf
    \EndFor
\EndWhile

\If{$|S| == 2$}
    \State \Return \textbf{true} if all creases of the remaining sectors in $S$ are identical
\EndIf

\State \Return \textbf{false}
\end{algorithmic}
\end{algorithm}

Global flat foldability—determining whether the full crease pattern admits a valid folded state without self-intersection—is historically verified via a continuous 'pointwise' definition. To make this computationally viable, \citet{FlatFolder_OSME2024} introduced a practical 'facewise' formulation that replaces this continuous check with a finite constraint-satisfaction problem over pairs of overlapping convex faces. In our Python reimplementation, initial face-pair assignments are derived from the mountain--valley crease labels and propagated through pre-computed implication tables. The remaining unassigned variables are partitioned into connected components, each solved independently via depth-first backtracking with constraint propagation.

\section{Packing}

\subsection{Grid Size Initialization Heuristic}
  \label{app:grid_size_heuristic}
  Before launching the backtracking packing search, we initialize the grid size using a heuristic lower bound derived from circle-packing theory. This heuristic estimates the required grid area by summing the area contributions $a_i$ of each stick $i$ in the stick figure, based on its type (flaps vs.\ rivers) and connectivity:
  \begin{equation}
      G_{\text{init}} = \max \left( \left\lceil \sqrt{ \sum_i a_i } \right\rceil, \, D \right)
  \end{equation}
  where $D$ is the diameter of the stick figure tree (the longest path in the tree weighted by stick length), and the individual area contributions $a_i$ are defined as follows:
  \begin{itemize}
      \item \textbf{Four Largest Flaps:} For the four longest flaps in the tree, we assume they are packed as quarter circles, contributing:
      \begin{equation}
          a_i = l_i^2
      \end{equation}
      where $l_i$ is the length of the flap.
      \item \textbf{Rivers with Only Flap Neighbors:} For a river of length $k_i$ where all adjacent sticks at its endpoints (excluding itself) are flaps, the area contribution is:
      \begin{equation}
          a_i = \max(k_i \cdot m_i, \, k_i^2)
      \end{equation}
      where $m_i$ is the maximum length among all flaps attached to the river's endpoints.
      \item \textbf{Rivers with Non-Flap Neighbors:} For a river that has at least one non-flap neighbor at its endpoints (e.g., another river), the area contribution is:
      \begin{equation}
          a_i = k_i^2
      \end{equation}
      \item \textbf{Remaining Sticks:} Any other sticks not covered by the rules above (e.g., additional flaps beyond the four largest) default to a conservative area contribution of:
      \begin{equation}
          a_i = 2 l_i^2
      \end{equation}
  \end{itemize}
  For symmetric stick figures, if the resulting $G_{\text{init}}$ is odd, it is incremented by $1$ to ensure an even grid size, maintaining symmetry during packing.

\begin{figure*}[t]
    \centering
    \includegraphics[width=\textwidth]{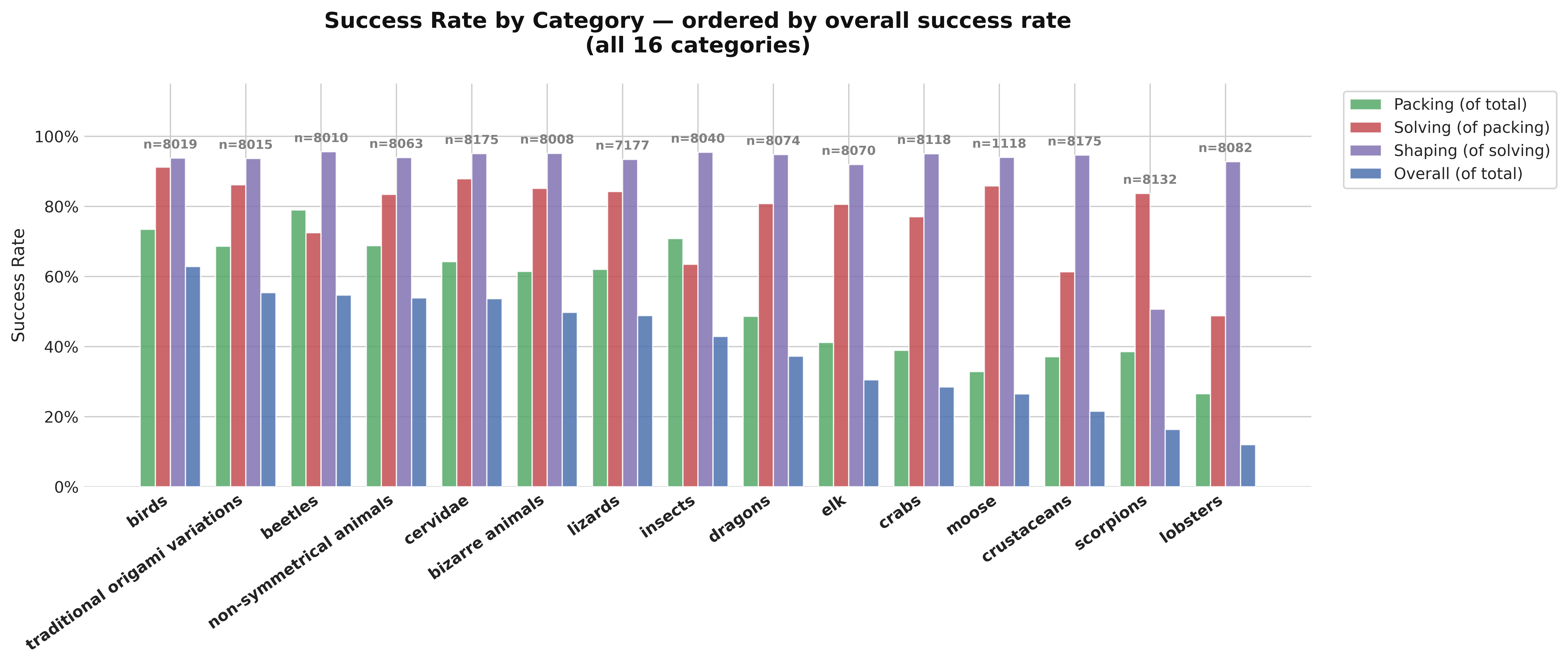}
    \caption{Success rate by category, ordered by overall success rate. The chart shows the packing, solving, shaping, and overall success rates for categories with at least 10 samples.}
    \label{fig:sr_by_category}
\end{figure*}

\section{Solving}

    

\begin{enumerate}
    \item Crease construction:
    \begin{enumerate}
        \item Hinge and ridge creases are provided in the packing crease pattern
        \item Pleat creases are constructed
    \end{enumerate}
    \item Mountain valley assignment:
    \begin{enumerate}
        \item Pleats and then ridge creases are assigned deterministically    \item A greedy algorithm is used to assign hinges and reassign pleats until a globally flat foldable solution is found
    \end{enumerate}
\end{enumerate}

As a reminder, a valid packing solution satisfies the following: 
\begin{itemize}
\item Each flap or river is packed to the paper with hinge and ridge creases.
\item The rivers and flaps occupy the entire paper. 
\item The hinges may be connected to each other. Each connected component of hinges is associated with a node in the tree.
\end{itemize}

Note that step 1 and step 2a are essentially deterministic and do not involve decision making. Step 2b on the other hand involves multiple assignment options for each hinge crease and is therefore combinatorial in the number of hinges. Nevertheless, we found that a greedy decision making algorithm is remarkably efficient as we now describe. 

\subsection{Deterministic steps}

\subsubsection{Pleat construction}
\label{subsub:pleats_con}
The axial pleats represent the fundamental grid of the box-pleated model. Their construction and initial assignment follow a rigid geometric logic before the hinge solver refines them. Pleats are generated by filtering a dense grid of axial candidates in five steps.


\begin{enumerate}
    \item  A dense, continuous orthogonal grid is globally mapped across the entire sheet.
    \item The continuous grid lines are split into discrete candidate segments at every intersection with previously defined hinges and ridges.
    \item The system applies strict structural heuristics, retaining only those segments that are strictly perpendicular to an intersecting hinge, or those that pass through an $X$-shape vertex (the precise junction of four intersecting ridges).
    \item Segments intersecting the outer border of the paper are selectively re-introduced, provided they share a common vertex with both a ridge and an actively established pleat.
    \item Finally, an iterative pass evaluates all remaining unoccupied grid coordinates, injecting structurally valid, non-intersecting pleat segments to guarantee comprehensive axial coverage and a fully contiguous fundamental grid.
\end{enumerate}


\subsection{Interleaving Assignment}
Initial fold orientations are assigned to pleats using a graph-based interleaving strategy:
\begin{itemize}
    \item Pleats are grouped into \textbf{paths} of connected segments.
    \item Paths are sorted by their distance from the origin.
    \item A Breadth-First Search (BFS) is employed to assign fold orientations. Adjacent paths separated by a single grid unit are assigned alternating Mountain and Valley orientations.
\end{itemize}

\subsection{Ridge Construction and Assignment}
Ridges are the diagonal creases ($45^{\circ}$) that reconcile the axial grid with the flap/river geometry.

\textbf{Anchor Folds}
Ridge assignment begins with "anchor" points where the orientation is forced or highly constrained:
\begin{itemize}
    \item \textbf{Y-Shape Vertices}: Vertices where two perpendicular ridges meet a single pleat tip. All three creases in a Y-shape typically share the same fold orientation, allowing the ridge fold to be "anchored" to the pleat's fold.
    \item \textbf{Border Anchors}: Ridges intersecting the border that lack Y-shape context are anchored using neighbor heuristics or arbitrarily assigned to Mountain.
\end{itemize}

\textbf{Propagation Logic.}
Once anchors are established, folds propagate along connected paths using geometric consistency rules:
\begin{itemize}
    \item \textbf{X-Shape}: Four ridges meeting at a point share the same orientation.
    \item \textbf{Parallel}: Adjacent parallel ridges must have opposite fold orientations (flipped).
    \item \textbf{Perpendicular}: Perpendicular ridges at an intersection share the same fold orientation.
\end{itemize}

\section{Shaping}
\label{sec:shaping_supp}
\subsection{Orchestrating simple fold}
For orchestrating the simple fold tool, we develop an algorithm which converts an already generated stick figure design into a series of simple folds. When applied to the base crease pattern, these simple folds shape the model into 3D in a way resembling the stick figure. This shaping algorithm generates simple folds for each flap and river in a breadth-first-search manner starting from the stick figure's root. When computing shaping for a stick $s_{\textup{child}}$, we first determine the folded state of the paper after applying simple folds to shape the BFS parent of $s_{\textup{child}}$, $s_{\textup{parent}}$. This is encoded by a \textit{3-frame} orientation matrix $\{v_1, v_2, v_3\} = F_{\textup{init}} \in \mathbb{R}^{3 \times 3}$ where the first row points in the same direction as the shaped parent stick and the remaining rows are determined recursively by the starting orientation of the stick figure root. The goal is now to produce a rotation matrix $R$ which, when multiplying $F_{\textup{init}}$, produces a 3-frame $F_{\textup{final}}$ whose first row points in the direction of $s_{\textup{child}}$. However, this rotation matrix $R$ must also be physically realizable as a simple fold. This imposes the additional constraint that the rotation axis of $R$ must lie in the plane determined by the span of $\{v_1, v_2\}$. Note: this plane is the initial paper plane before shaping $s_{\textup{child}}$ i.e. the plane containing the flap before shaping. This additional constraint uniquely determines $R$ and the final 3-frame $R F_{\textup{init}} = F_{\textup{final}}$. $R$ also gives us the argument to the simple fold shaping tool we need: the rotation axis i.e. the simple fold line.

\subsection{Narrowing Templates for the Clip Pattern Algorithm}
\label{app:clip_patterns}

The clip pattern algorithm relies on a library of pre-defined 2D crease templates, which are normalized to the orthogonal grid and oriented to produce a desired effect when projected onto the folded segment. In this section, we present a family of narrowing templates we developed for shaping.

The general structure of a narrowing template consists of a base adapter and a narrowing pleat. The role of the base adapter is to divert narrowing pleats such that the rest of the model is not impacted by the narrowing. This is achieved by having a crease pattern where the starting edge does not have vertices inside it, yet the pattern remains flat-foldable and achieves the narrowing effect.

\begin{figure}[htbp]
    \centering
    \captionsetup[subfigure]{justification=centering} 
    \begin{subfigure}[b]{0.48\textwidth}
        \centering
        \includegraphics[width=\textwidth]{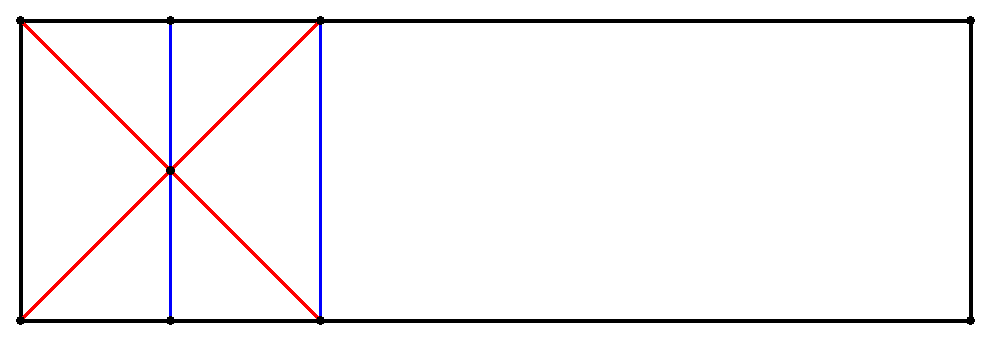}
        \caption{1x Symmetric Flap Narrowing (Structural Base)}
    \end{subfigure}
    \hfill
    \begin{subfigure}[b]{0.48\textwidth}
        \centering
        \includegraphics[width=\textwidth]{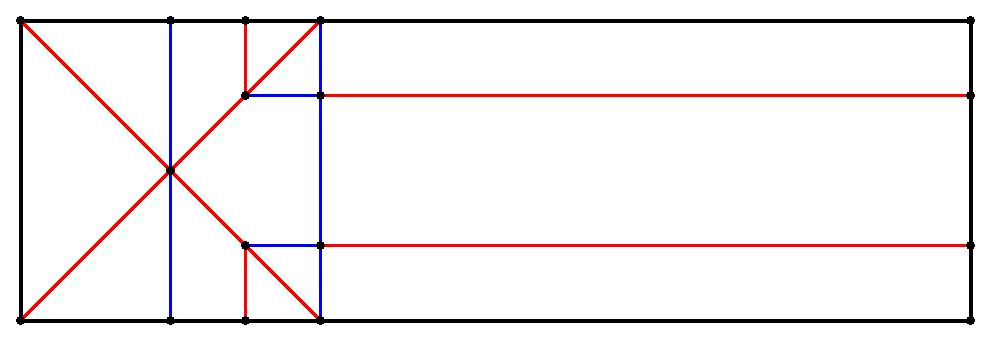}
        \caption{2x Symmetric Flap Narrowing}
    \end{subfigure}
    
    \vspace{1.5em}
    
    \begin{subfigure}[b]{0.48\textwidth}
        \centering
        \includegraphics[width=\textwidth]{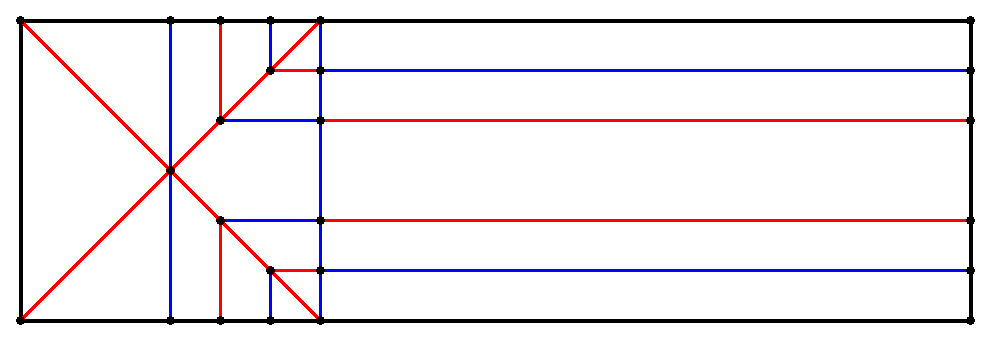}
        \caption{3x Symmetric Flap Narrowing}
    \end{subfigure}
    \hfill
    \begin{subfigure}[b]{0.48\textwidth}
        \centering
        \includegraphics[width=\textwidth]{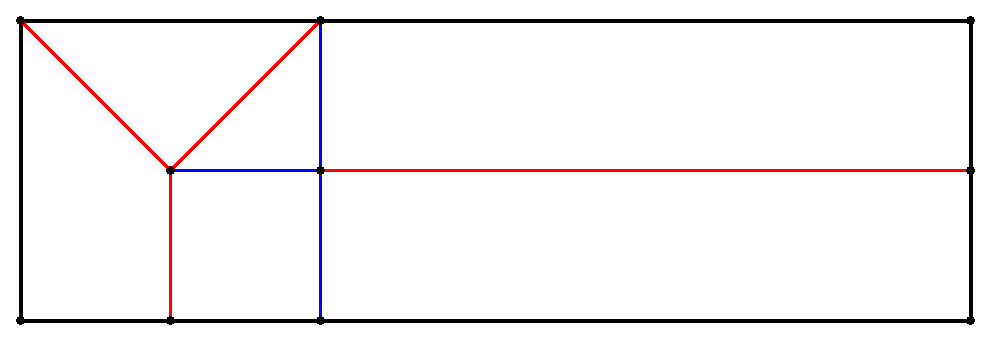}
        \caption{2x Asymmetric Flap Narrowing (Simplified Base)}
    \end{subfigure}
    \caption{A grid comparison of flap narrowing templates. (a--c) Symmetric templates demonstrate how dynamically adjusting the internal pleat network parametrically reduces the width of a structural segment. (d) The asymmetric template simplifies the adapter block to shift mass to one side, extending outward until geometrically clipped by the layer's hull.}
    \label{fig:flap_templates}
\end{figure}

\begin{figure}[htbp]
    \centering
    \captionsetup[subfigure]{justification=centering} 
    \begin{subfigure}[b]{0.48\textwidth}
        \centering
        \includegraphics[width=\textwidth]{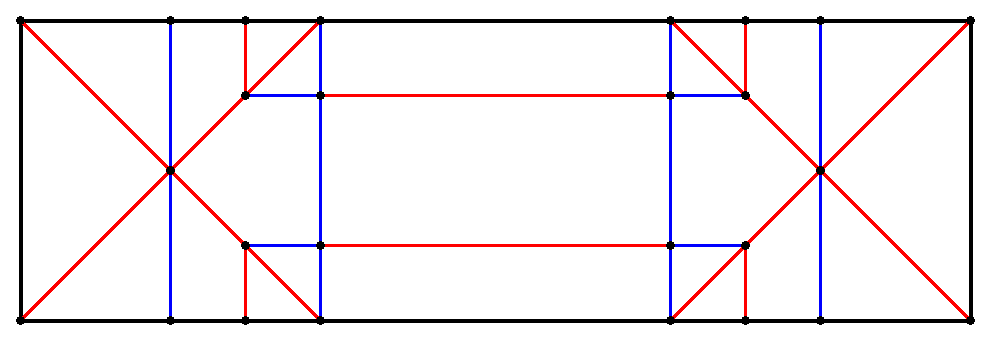}
        \caption{2x Symmetric River Narrowing}
    \end{subfigure}
    \hfill
    \begin{subfigure}[b]{0.48\textwidth}
        \centering
        \includegraphics[width=\textwidth]{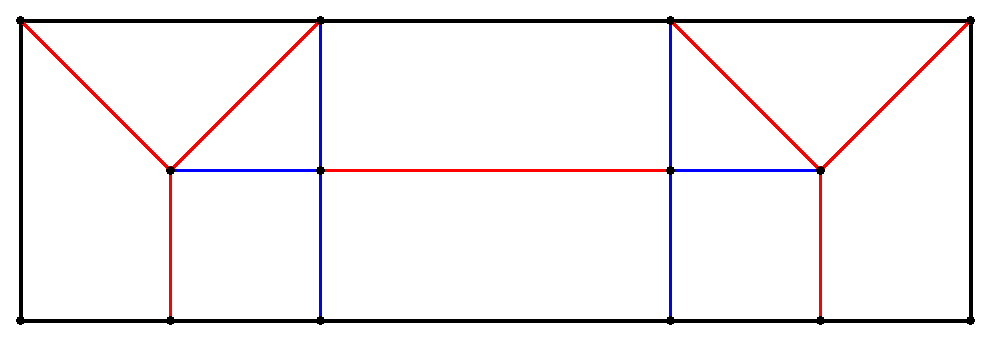}
        \caption{2x Asymmetric River Narrowing (Simplified Base)}
    \end{subfigure}
    \caption{Crease patterns for narrowing rivers. Mirrored adapter blocks at both ends safely reintegrate the narrowed segment with the surrounding orthogonal grid. The asymmetric variation (b) provides a more crease-efficient transition compared to the symmetric variation (a).}
    \label{fig:river_templates}
\end{figure}

For completeness, we describe and display the symmetrical and asymmetrical templates we developed, as well as how they look when applied to flaps versus rivers (see Figure \ref{fig:flap_templates} and Figure \ref{fig:river_templates}).

\textbf{Symmetrical and Asymmetrical.}\quad A folded flap or river can be narrowed in two ways. Symmetric templates pinch both the left and right edges inwards toward the centre axis, which is ideal for preserving bilateral symmetry in appendages like insect legs. Asymmetric templates, conversely, fold only one edge over and tuck it inside the other, effectively halving the width of the flap while shifting its mass to one side. Technically, a symmetric narrowing template can be used asymmetrically if only one side of the paper is folded. However, in the case of a 2x fold, creating a dedicated asymmetric template allows for a base adapter with fewer creases. For some models, asymmetric narrowing may create a desired visual effect, such as shifting the origin of a leg more to one side.

\textbf{Flaps vs. Rivers.}\quad A segment's position within the model determines its boundary conditions. Because flaps act as terminal appendages, they require only a single adapter at their base; the narrowed pleats extend outward until they are clipped by the algorithm. Conversely, rivers act as internal structural bridges (e.g., an animal's neck). To maintain connectivity between parts, river templates require two adapters: an initial one to narrow the paper, and a mirrored one at the opposite end to return the paper to the standard grid width. Consequently, the overall length of the river acts as a dynamic parameter of the template, which must explicitly account for the physical footprint of both base adapters. If a designated river segment is too short to accommodate these adapters, it cannot be narrowed using this method.

\section{Folding}
\label{sec:folding_supp}
To compute the spatial configuration, the algorithm constructs a face-adjacency graph and executes a breadth-first traversal. For each newly visited adjacent face, we compute a global $4 \times 4$ affine transformation matrix. This is constructed by translating the shared edge to the origin, applying a rotation matrix around the edge's axis by the specified fold angle, applying the inverse translation, and composing the result with the parent face's transformation matrix.

\begin{figure}[htbp]
    \centering
    \includegraphics[width=0.7\textwidth]{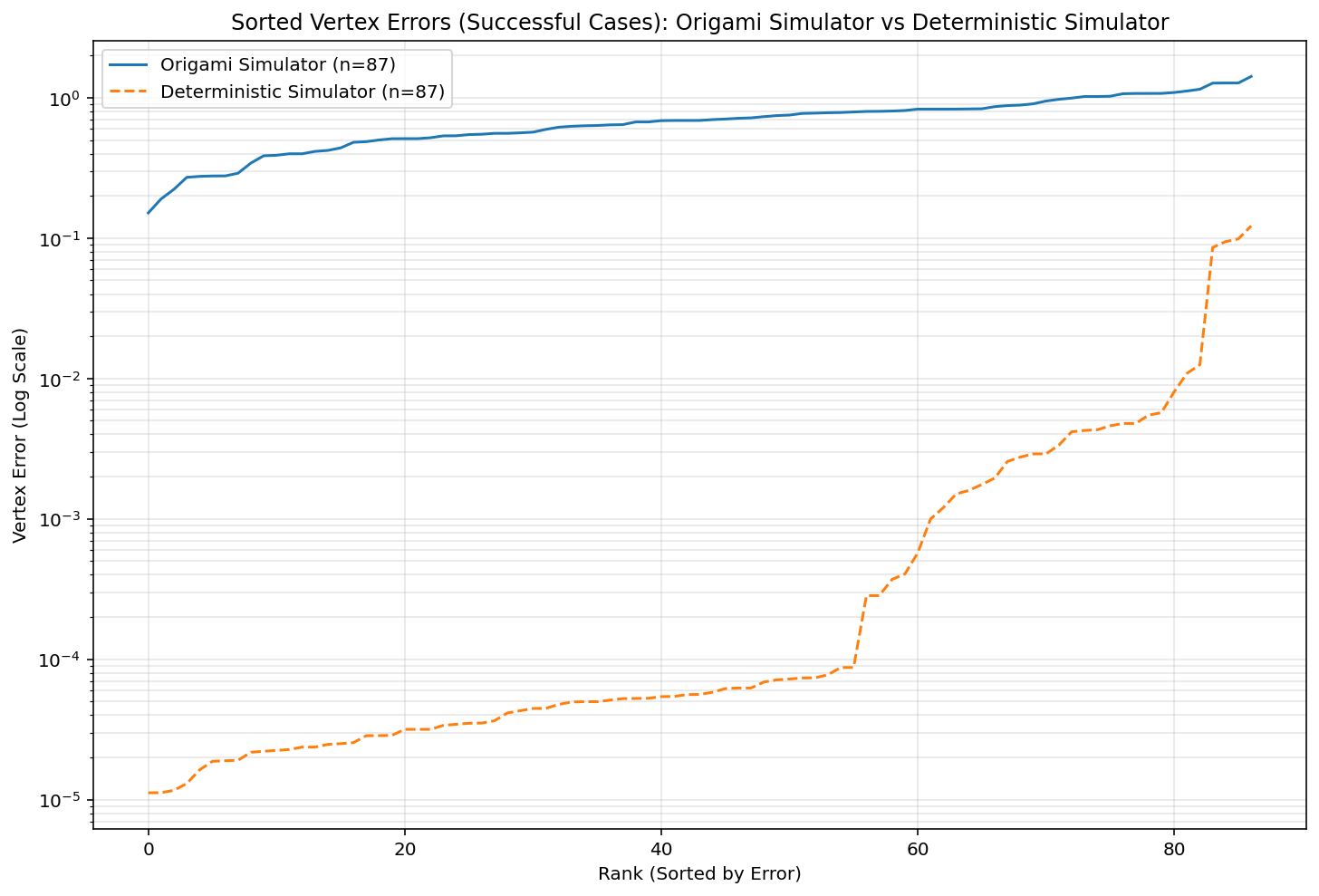}
    \caption{\textbf{Numerical Accuracy Comparison of Folding Solvers.} We plot the sorted vertex reconstruction errors on a logarithmic scale for 87 complex box-pleated crease patterns containing thousands of creases. The comparison evaluates our deterministic folding solver (orange dashed line) against the baseline mass-spring dynamic model from Origami Simulator~\cite{ghassaei2018fast} (blue solid line). Across all test cases, the proposed deterministic solver consistently achieves lower geometric errors, demonstrating up to a five-order-of-magnitude precision improvement over the physics-based baseline.}
    \label{fig:folder_comparison}
\end{figure}

In \cref{fig:folder_comparison}, we evaluate the numerical precision of our deterministic folding engine against the mass-spring dynamic model implemented in Origami Simulator~\citep{ghassaei2018fast} across 87 complex crease patterns containing up to several thousand creases. The figure plots the sorted vertex reconstruction errors for both approaches on a logarithmic scale. As demonstrated, the deterministic folding solver consistently yields significantly lower geometric errors than the mass-spring baseline across all test cases, achieving an error reduction of up to five orders of magnitude (with vertex errors falling as low as $10^{-5}$ compared to Origami Simulator's $10^{-1}$ baseline).

\section{VLM feedback}
\label{apx:vlm_feedback_prompt}
As mentioned, our VLM evaluation pipeline operates in two distinct modes: single model evaluation mode and comparison judge mode. Here we provide the prompt we use for Gemini 3 Flash.

\vspace{1em}
\begin{tcolorbox}[
    colback=gray!5!white, 
    colframe=gray!75!black, 
    arc=4pt, 
    boxrule=0.5pt,
    title=Single Model Evaluation Mode Prompt,
    fontupper=\small, 
    fonttitle=\bfseries,
    coltitle=white
]
\{\{image\_views\}\}\\

\# Role

You are a Rigorous Scientific Origami Judge. Your primary objective is to evaluate a 3D-simulated fold based on its anatomical accuracy and biological proportions. While technical folding skill is required, anatomical fidelity is paramount. Color, shading, and surface patterns are to be entirely ignored, as the model is folded from a single sheet of monochrome paper; your evaluation must focus exclusively on form, structure, geometry and aesthetics. You must explicitly identify the subject's primary anatomical segments. Any model that presents the body as a single, undifferentiated mass without clear transitions between major regions (such as Head vs. Thorax) will be disqualified. \\

\# Input Data

* **Target Object:** \{\{text\_desc\}\}

* **Visual Data:** Multiple 2D snapshots of the final 3D folded state from different angles.\\

\# Task

Analyze the provided images to determine how well the folded mesh corresponds to the semantic concept of a "\{\{text\_desc\}\}". You must look for key defining geometric features (e.g., wings for a bird, legs for a table, symmetry where applicable).\\

\# Analysis Steps

1.  **Feature Identification:** List the visual features visible in the images.

2.  **Comparison:** Compare these features against the real-world counterpart shape of a \{\{text\_desc\}\} (using internal knowledge or provided references). You must explicitly verify:

\hspace*{1em}*   Appendage Count: (Primary) Does it have the correct number of appendages? (e.g., A beetle must have 6 distinct legs and 2 antennae).
    
\hspace*{1em}*   Topology: (Secondary) Do appendages emerge from the correct anatomical region (shoulders/hips)? (e.g., insect legs must emerge from the thorax, not the abdomen; wings must emerge from the shoulder/upper back)
    
\hspace*{1em}*   Proportionality: (Tertiary) Are the features scaled correctly relative to the real creature? (e.g., Are the antennae thread-like or thick blocks? Are the legs jointed or simple triangles?)
    
\hspace*{1em}*   Differentiation: (Quaternary) Is there a clear geometric change between the head, neck, and body?

\hspace*{1em}*   Aesthetic Refinement: (Quinary) Are the folds sharp and the symmetry balanced?
    
3.  **Flaw Detection:** Identify geometric and structural failures that compromise the model's scientific validity. Critically evaluate for structural congestion, where tangled or overlapping flaps obscure the intended anatomy; anatomical undifferentiation, where the body mass lacks distinct segmentation from the limbs. Additionally, check for standard failures such as crumbling, intersecting meshes, or severe asymmetry.

4.  **Scoring:** Assign a score based on the rubric below. Do not penalize the model for being monochromatic.\\

\# Scoring Rubric (0-10)

* **0 (Unrecognizable):** The mesh looks like a crumpled ball, flat paper, or random noise. No features of a \{\{text\_desc\}\} are present.

* **2 (Abstract/Vague):** Vaguely resembles the *mass* of the object, but lacks distinct limbs/parts. (e.g., it is "bird-shaped" but has no distinct wings or head).

* **4 (Poor Correspondence):** The object is recognizable as an attempt at the target subject, but possesses major structural deformities—such as a lack of distinct differentiation between the body mass and the appendages—or missing key parts (e.g., pincers).

* **6 (Fair Correspondence):** Recognizable as the general class of object, but contains significant anatomical errors, such as an incorrect number of major limbs (e.g., a 4-legged insect) or severely skewed proportions (e.g., overly thick antennae).

* **8 (Good Correspondence):** Strong resemblance and clean folding. Key features are present and in the correct numbers (e.g., 6 legs), but their shapes may be slightly abstracted or blocky.

* **10 (Perfect Correspondence):** A pristine fold that is anatomically correct. It has the exact count of limbs/features relative to the real subject, and their shapes and proportions are highly accurate representations.\\

\# Note

The images are taken from different views, so it is possible that the origami looks different or even not recognizable from some views.

Thus, you should pay more attention to the key views that most clearly show correspondence.\\

\# Output Format

Provide your reasoning first, then the final score as a number inside angle brackets, e.g. <5>.
\end{tcolorbox}

\vspace{1em}
\begin{tcolorbox}[
    colback=gray!5!white, 
    colframe=gray!75!black, 
    arc=4pt, 
    boxrule=0.5pt,
    title=Comparison Judge Mode Prompt,
    fontupper=\small, 
    fonttitle=\bfseries,
    coltitle=white,
    breakable,
]
Reference Images of a "\{\{text\_desc\}\}":

\{reference\_image\_views\}\\

Candidate Images:

\{image\_views\}\\

\# Role

You are a Rigorous Scientific Origami Judge. Your primary objective is to compare a **Candidate** origami against a **Reference** origami of the same target subject and determine which one is a better representation. The reference is NOT necessarily high quality — it is simply a baseline for comparison. Color, shading, and surface patterns are to be entirely ignored, as the models are folded from a single sheet of monochrome paper; your evaluation must focus exclusively on form, structure, geometry and aesthetics.\\

\# Input Data

* **Target Object:** \{\{text\_desc\}\}

* **Reference Images:** Images of a reference "\{\{text\_desc\}\}" origami. This serves as a baseline for comparison — it may be of any quality level.

* **Candidate Images:** Multiple 2D snapshots of the candidate 3D folded state from different angles.\\

\# Task

Compare the **Candidate Images** against the **Reference Images** and determine whether the candidate origami is a **better or worse** representation of a "\{\{text\_desc\}\}" than the reference. You must assess both origami on their structural accuracy, proportional fidelity, geometric quality, and overall recognizability, then judge which one is superior.\\

\# Analysis Steps

1.  **Reference and Candidate Analysis:** Study the reference images, Compare reference's and candidate's features against the real-world counterpart shape of a \{\{text\_desc\}\} (using internal knowledge or provided references). You must explicitly verify:

\hspace*{1em}*   Appendage Count: (Primary) Does it have the correct number of appendages? (e.g., A beetle must have 6 distinct legs and 2 antennae).
    
\hspace*{1em}*   Topology: (Secondary) Do appendages emerge from the correct anatomical region (shoulders/hips)? (e.g., insect legs must emerge from the thorax, not the abdomen; wings must emerge from the shoulder/upper back)
    
\hspace*{1em}*   Proportionality: (Tertiary) Are the features scaled correctly relative to the real creature? (e.g., Are the antennae thread-like or thick blocks? Are the legs jointed or simple triangles?)
    
\hspace*{1em}*   Differentiation: (Quaternary) Is there a clear geometric change between the head, neck, and body?
    
\hspace*{1em}*   Aesthetic Refinement: (Quinary) Are the folds sharp and the symmetry balanced?.

2.  **Comparative Assessment:** Directly compare the candidate against the reference on each dimension:
\hspace*{1em}*   **Recognizability:** Which origami is more immediately recognizable as a "\{\{text\_desc\}\}"?
    
\hspace*{1em}*   **Structural Accuracy:** Which has more correct and well-defined structural features (appendages, body segments, distinctive shapes)?
    
\hspace*{1em}*   **Proportional Fidelity:** Which has proportions closer to the real-world "\{\{text\_desc\}\}"?
    
\hspace*{1em}*   **Geometric Quality:** Which has cleaner, sharper folds and better overall craftsmanship?
    
\hspace*{1em}*   **Aesthetic Quality:** Which is more aesthetically pleasing as an origami piece?

3.  **Overall Judgment:** Based on the above comparisons, determine whether the candidate is significantly better, somewhat better, roughly equal, somewhat worse, or significantly worse than the reference.

4.  **Scoring:** Assign a score based on the rubric below.\\

\# Scoring Rubric (0-10)

* **0 (Candidate is drastically worse):** The candidate is a crumpled ball or completely unrecognizable, while the reference is a reasonable origami.

* **2 (Candidate is notably worse):** The reference is clearly a better representation. The candidate has major structural flaws or missing features that the reference does not.

* **4 (Candidate is slightly worse):** The reference is marginally better. Both are comparable but the reference has an edge in some dimensions.

* **5 (Roughly equal):** Both origami are of similar quality. Neither is clearly better than the other.

* **6 (Candidate is slightly better):** The candidate is marginally better. Both are comparable but the candidate has an edge in some dimensions.

* **8 (Candidate is notably better):** The candidate is clearly a better representation. It has better structure, proportions, or recognizability than the reference.

* **10 (Candidate is drastically better):** The candidate is an excellent origami while the reference is poor or unrecognizable.\\

\# Note
The candidate and reference images may be taken from different viewing angles. Focus on comparing structural features and overall form rather than exact angle-by-angle correspondence.\\

\# Output Format
Provide your reasoning first, then the final score as a number inside angle brackets, e.g. <7> or <3>.
\end{tcolorbox}

\subsection{VLM prompt baselines}
\vspace{1em}
\begin{tcolorbox}[
    colback=gray!5!white, 
    colframe=gray!75!black, 
    arc=4pt, 
    boxrule=0.5pt,
    title=Binary,
    fonttitle=\bfseries,
    coltitle=white
]
Candidate Images:

\{image\_views\}\\

\# Input Data \\
* Target Object: \{\{text\_desc\}\} \\
* Visual Data: Multiple 2D snapshots of the final 3D folded state from different angles. \\

\# Task \\
Evaluate whether the 3D-folded model in the images successfully represents a "\{\{text\_desc\}\}". If it is a recognizable, successful origami version of a "\{\{text\_desc\}\}", output <true>. If it is a failure, unrecognizable, a crumpled mass of paper, or completely incorrect, output <false>.\\

Provide your reasoning first, and then on a new line output either <true> or <false>.
\end{tcolorbox}
\vspace{1em}

\vspace{1em}
\begin{tcolorbox}[
    colback=gray!5!white, 
    colframe=gray!75!black, 
    arc=4pt, 
    boxrule=0.5pt,
    title=Score,
    fonttitle=\bfseries,
    coltitle=white
]
Candidate Images:

\{image\_views\}\\

\# Input Data \\
* Target Object: \{\{text\_desc\}\} \\
* Visual Data: Multiple 2D snapshots of the final 3D folded state from different angles. \\

\# Task \\
Evaluate whether the 3D-folded model in the images represents a "\{\{text\_desc\}\}". Rate the success and quality of the origami piece on a scale from 0 to 10, where 0 is a complete failure/unrecognizable shape/crumpled paper, and 10 is a perfectly recognizable, clean origami shape of "\{\{text\_desc\}\}".

Provide your reasoning first, and then on a new line output the score inside angle brackets, e.g. <7>.
\end{tcolorbox}
\vspace{1em}

\vspace{1em}
\begin{tcolorbox}[
    colback=gray!5!white, 
    colframe=gray!75!black, 
    arc=4pt, 
    boxrule=0.5pt,
    title=Rubrics V0,
    fonttitle=\bfseries,
    coltitle=white
]
\# Role\\
You are a master Origami Judge and geometric analyst. You are evaluating a 3D simulation of an origami folding process executed by an AI agent.\\

Candidate Images:

\{image\_views\}\\

\# Input Data \\
* Target Object: \{\{text\_desc\}\} \\
* Visual Data: Multiple 2D snapshots of the final 3D folded state from different angles. \\

\# Task \\
Analyze the provided images to determine how well the folded mesh corresponds to the semantic concept of a "\{\{text\_desc\}\}". You must look for key defining geometric features (e.g., wings for a bird, legs for a table, symmetry where applicable).

\# Analysis Steps\\
1.  **Feature Identification:** List the visual features visible in the images.\\
2.  **Comparison:** Compare these features against the canonical shape of a \{\{text\_desc\}\}.\\
3.  **Flaw Detection:** Identify geometric failures (e.g., crumbling, intersecting meshes, lack of symmetry, unrecognizable silhouette).\\
4.  **Scoring:** Assign a score based on the rubric below.\\

\# Scoring Rubric (0-10)\\
* **0 (Unrecognizable):** The mesh looks like a crumpled ball, flat paper, or random noise. No features of a \{\{text\_desc\}\} are present.\\
* **2 (Abstract/Vague):** Vaguely resembles the *mass* of the object, but lacks distinct limbs/parts. (e.g., it is "bird-shaped" but has no distinct wings or head).\\
* **4 (Poor Correspondence):** The object is recognizable as an attempt at a \{\{text\_desc\}\}, but has major structural deformities, severe asymmetry, or missing key parts.\\
* **6 (Fair Correspondence):** Clearly recognizable as a \{\{text\_desc\}\}. Major features are present but may be messy, slightly crumpled, or have poor proportions.\\
* **8 (Good Correspondence):** Strong resemblance. Key features are distinct and geometric. Minor folding artifacts or slight asymmetry only.\\
* **10 (Perfect Correspondence):** A pristine fold. Distinct, clean geometry that perfectly matches the ideal silhouette of a \{\{text\_desc\}\} from all angles.\\

\# Note\\
The images are taken from different views, so it is possible that the origami looks different or even not recognizable from some views. Thus, you should pay more attention to the key views that most clearly show correspondence. And, most simulation samples should be flat-foldable, so it is normal to have "flat" or 2D views, do not penalize for not being 3D.\\

\# Output Format\\
Provide your reasoning first, then the final score as a number inside angle brackets, e.g. <5>.
\end{tcolorbox}
\vspace{1em}
\end{document}